\documentclass[10pt,journal,compsoc]{IEEEtran}
\ifCLASSINFOpdf
\else
\fi

\usepackage[belowskip=-15pt,aboveskip=0pt]{caption}
\usepackage{amssymb}
\usepackage{algorithmic}
\usepackage{float}
\usepackage[utf8]{inputenc}
\usepackage{lipsum}
\usepackage{ifmtarg}
\usepackage{array}
\usepackage{amsmath}

\usepackage{color}
\usepackage{tabularx}
\usepackage[colorlinks=true, allcolors=green]{hyperref}
\usepackage{adjustbox}
\usepackage{epstopdf}
\usepackage[ruled,linesnumbered,vlined,commentsnumbered]{algorithm2e}
\usepackage{cite}
\usepackage{booktabs}
\usepackage{multirow}
\usepackage{longtable}

\usepackage{blindtext}

\usepackage{caption}
\usepackage{subcaption}
\graphicspath{{figures/}}

\begin{document}

\title{A Review of Generalized Zero-Shot Learning Methods}

\author{Farhad~Pourpanah,~\IEEEmembership{Member,~IEEE,}
         Moloud~Abdar,
         Yuxuan~Luo,
         Xinlei~Zhou,
         Ran~Wang,~\IEEEmembership{Member,~IEEE,}
         Chee~Peng~Lim,
       Xi-Zhao~Wang,~\IEEEmembership{Fellow,~IEEE}
       and~Q.~M.~Jonathan~Wu,~\IEEEmembership{Senior Member,~IEEE}

\IEEEcompsocitemizethanks{\IEEEcompsocthanksitem This work is partially supported by the National Natural Science Foundation of China (Grant nos. 62176160, 61732011 and 61976141), the Guangdong Basic and Applied Basic Research Foundation (Grant 2022A1515010791), the Natural Science Foundation of Shenzhen (University Stability Support Program no. 20200804193857002), and the Interdisciplinary Innovation Team of SZU (\textit{Corresponding authors: Ran Wang \& Xi-Zhao Wang}).\protect 
\IEEEcompsocthanksitem F. Pourpanah and Q. M. J. Wu are with the Centre for Computer Vision and Deep Learning, Department of Electrical and Computer Engineering, University of Windsor, Windsor, ON N9B 3P4, Canada (e-mails: farhad.086@gmail.com and jwu@uwindsor.ca).\protect
\IEEEcompsocthanksitem M. Abdar and C. P. Lim are with the Institute for Intelligent Systems Research and Innovation (IISRI), Deakin University, Australia (e-mails: m.abdar1987@gmail.com \& chee.lim@deakin.edu.au).\protect
\IEEEcompsocthanksitem Y. Luo is with the Department of Computer Science, City University of Hong Kong, Hong Kong SAR, China (e-mail: yuxuanluo4-c@my.cityu.edu.hk).\protect
\IEEEcompsocthanksitem R. Wang is with the College of Mathematics and Statistics, Shenzhen Key Lab. of Advanced Machine Learning and Applications, Shenzhen University, Shenzhen 518060, China (e-mail: wangran@szu.edu.cn).\protect
\IEEEcompsocthanksitem  X. Zhou and X. Wang are with the College of Computer Science and Software Engineering, Guangdong Key Lab. of Intelligent Information Processing, Shenzhen University, Shenzhen 518060, China (e-mails: zhouxnli@163.com \& xizhaowang@ieee.org).
}
}


\IEEEtitleabstractindextext{
\begin{abstract}
Generalized zero-shot learning (GZSL) aims to train a model for classifying data samples under the condition that some output classes are unknown during supervised learning.
To address this challenging task, GZSL leverages semantic information of the seen (source) and unseen (target) classes to bridge the gap between both seen and unseen classes. Since its introduction, many GZSL models have been formulated.
In this review paper, we present a comprehensive review on GZSL. Firstly, we provide an overview of GZSL including the problems and challenges.
Then, we introduce a hierarchical categorization for the GZSL methods and discuss the representative methods in each category.
In addition, we discuss the available benchmark data sets and applications of GZSL, along with a discussion on the research gaps and directions for future investigations.

\end{abstract}
}

\maketitle

\IEEEdisplaynontitleabstractindextext

\IEEEpeerreviewmaketitle

\section{Introduction}
\label{Sec:intro}
\IEEEPARstart{W}{ith} recent advances in image processing and computer vision, deep learning (DL) models have achieved extensive popularity due to their capability for providing an end-to-end solution from feature extraction to classification.
{\color{black}
Despite their success, traditional DL models require training on a massive amount of labeled data for each class, along with a large number of samples.
In this respect, it is a challenging issue to collect large-scale labelled samples. }
As an example, ImageNet~\cite{imahenet2009deng}, which is a large data set, contains 14 million images with 21,814 classes in which many classes contain only few images.
In addition, standard DL models can only recognize samples belonging to the classes that have been seen during the training phase, and they are not able to handle samples from unseen classes~\cite{liu2018generalized}.
While in many real-world scenarios, there may not be a significant amount of labeled samples for all classes.
On one hand, fine-grained annotation of a large number of samples is laborious and it requires an expert domain knowledge.
On the other hand, many categories are lack of sufficient labeled samples, e.g., endangered birds, or being observed in progress, e.g., COVID-19, or not covered during training but appear in the test phase~\cite{wang2019survey,wang2020recent,kabir2020spinalnet,abdar2020review}.\par

\begin{figure}
    \includegraphics[width=0.49\textwidth]{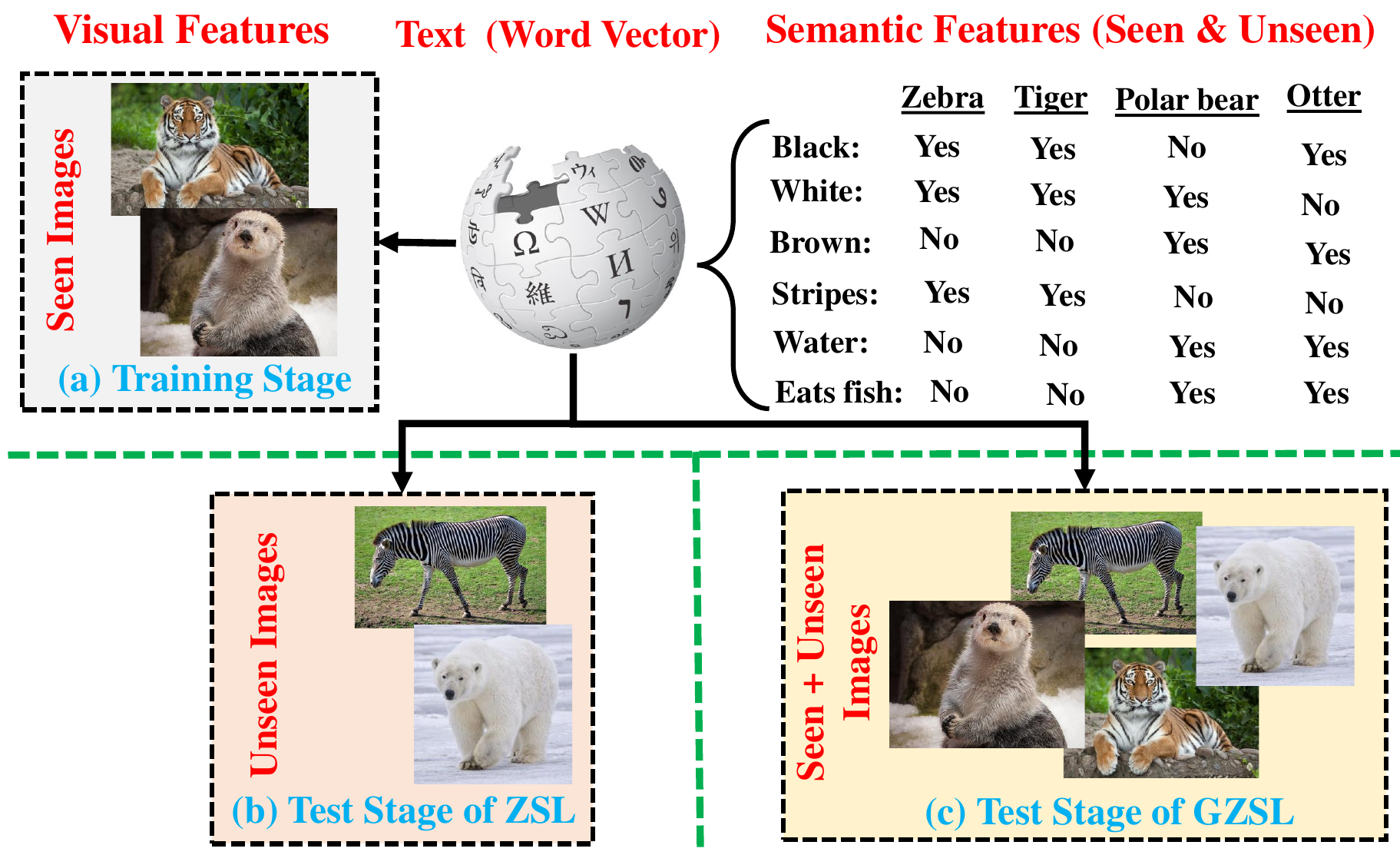}
    \caption{{\color{black}A schematic diagram of ZSL versus GZSL. Assume that the seen class contains samples of \textit{Otter} and \textit{Tiger}, while the unseen class contains samples of \textit{Polar bear} and \textit{Zebra}. (a) During the training phase, both GZSL and ZSL methods have access to the samples and semantic representations of the seen class. (b) During the test phase, ZSL can only recognize samples from the unseen class, while (c) GZSL is able to recognize samples from both seen and unseen classes.}}
    \label{fig:ZSLvsGZSL}
\end{figure}

{\color{black}Several techniques for various learning configurations have been developed. \emph{One-shot}~\cite{fei2006one} and \emph{few-shot}~\cite{scheirer2012toward} learning techniques can learn from classes with a few learning samples. These techniques use the knowledge obtained from data samples of other classes and formulate a classification model for handling classes with few samples. While the \textit{open set recognition} (OSR)~\cite{scheirer2013toward} techniques can identify whether a test sample belongs to an unseen class, they are not able to predict an exact class label.  Out-of-distribution~\cite{yang2021generalized} techniques attempt to identify test samples that are different from the training samples. However, none of the above-mentioned techniques can classify samples from unseen classes.
In contrast, human can recognize around 30,000 categories~\cite{biederman1987recognition}, in which we do not need to learn all these categories in advance. As an example, a child can easily recognize zebra, if he/she has seen horses previously, and have the knowledge that a zebra looks like a horse with black and white strips.  Zero-shot learning (ZSL)~\cite{larochelle2008zero,lampert2014attribute} techniques offer a good solution to address such challenge.}\par

ZSL aims to train a model that can classify objects of unseen classes (target domain) via transferring knowledge obtained from other seen classes (source domain) with the help of semantic information.
The semantic information embeds the names of both seen and unseen classes in high-dimensional vectors.
Semantic information can be manually defined attribute vectors ~\cite{lampert2009learning}, automatically extracted word vectors~\cite{socher2013zero}, context-based embedding~\cite{fu2018zero}, or their combinations~\cite{song2020generalizedd,zhang2020pseudo}.
In other words, ZSL uses semantic information to bridge the gap between the seen and unseen classes.
This learning paradigm can be compared to a human when recognizing a new object by measuring the likelihoods between its descriptions and the previously learned notions~\cite{romera2015embarrassingly}.
{\color{black}In conventional ZSL techniques, the test set only contains samples from the unseen classes, which is an unrealistic setting and it does not reflect the real-world recognition conditions.
In practice, data samples of the seen classes are more common than those from the unseen ones, and it is important to recognize samples from both classes simultaneously rather than classifying only data samples of the unseen classes.
This setting is called generalized zero-shot learning (GZSL)~\cite{chao2016empirical}. Indeed, GZSL is a pragmatic version of ZSL. The main motivation of GZSL is to imitate human recognition capabilities, which can recognize samples from both seen and unseen classes.}
Fig.~\ref{fig:ZSLvsGZSL} presents a schematic diagram of GZSL and ZSL.\par

Socher et al.~\cite{socher2013zero} was the first to introduce the concept of GZSL in 2013. An outlier detection method is integrated into their model to determine whether a given test sample belongs to the manifold of seen classes. If it is from a seen class, a standard classifier is used; otherwise a class is assigned to the image by computing the likelihood of being an unseen class. In the same year, Frome et al.~\cite{frome2013devise} attempted to learn semantic relationships between labels by leveraging the textual data and then mapping the images into the semantic embedding space. Norouzi et al.~\cite{norouzi2013zero} used a convex combination of the class label embedding vectors to map images into the semantic embedding space, in an attempt to recognize samples from both seen and unseen classes.
However, GZSL did not gain traction until 2016, when Chao et al.~\cite{chao2016empirical} empirically showed that the techniques under ZSL setting cannot perform well under the GZSL setting.
This is because ZSL is easy to overfit on the seen classes, i.e., classify test samples from unseen classes as a class from the seen classes.
Later, Xian et al.~\cite{xian2017zero,xian2018zero} and Liu et al.~\cite{liu2019generalized} obtained similar findings on image and web-scale video data with ZSL, respectively.
This is mainly because of the strong bias of the existing techniques towards the seen classes in which almost all test samples belonging to unseen classes are classified as one of the seen classes.
To alleviate this issue, Chao et al.~\cite{chao2016empirical} introduced an effective calibration technique, called calibrated stacking, to balance the trade-off between recognizing samples from the seen and unseen classes, which allows learning knowledge about the unseen classes.
Since then, the number of proposed techniques under the GZSL setting has been increased radically.\par

\subsection{Contributions}
\label{Sec:sec:cont}
Since its introduction, GZSL has attracted the attention of many researchers. Although several comprehensive reviews of ZSL models can be found in the literature~\cite{wang2019survey,xian2018zero,fu2018recent,rezaei2020zero}. {\color{black} The main differences between our review paper with previous ZSL survey literature~\cite{wang2019survey,xian2018zero,fu2018recent,rezaei2020zero} are as follows. In~\cite{wang2019survey}, the mainly focus is on ZSL, and only a few GZSL methods have been reviewed. In~\cite{xian2018zero}, the impacts of various ZSL and GZSL methods in different case studies are investigated. Several SOTA ZSL and GZSL methods have been selected and evaluated using different data sets. However, the work~\cite{xian2018zero} is more focused on empirical research rather than a review paper on ZSL and GZSL methods. The study in~\cite{fu2018recent} is focused on ZSL, with only a brief discussion (a few paragraphs) on GZSL. Rezaei and Shahidi~\cite{rezaei2020zero} studied the importance of ZSL methods for COVID-19 diagnosis (medical application). Unlike the aforementioned review paper, in this manuscript, we focus on GZSL, rather than ZSL, methods. }
However, none of them include an in-depth survey and analysis of GZSL.
To fill this gap, we aim to provide a comprehensive review of GZSL in this paper, including the problem formulation, challenging issues, hierarchical categorization, and applications.

We review published articles, conference papers, book chapters, and high-quality preprints (i.e. arXiv) related to GZSL commencing from its popularity in 2016 till early 2021.
However, we may miss some of the recently published studies, which is not avoidable. In summary, the main contributions of this review paper include:
\begin{itemize}
\item comprehensive review of the GZSL methods, to the best of our knowledge, this is the first paper that attempts to provide an in-depth analysis of the GZSL methods;

\item hierarchical categorization of the GZSL methods along with their corresponding representative models and real-world applications;

\item elucidation on the main research gaps and suggestions for future research directions.
\end{itemize}

\vspace*{-0.25cm}
{\color{black}
\subsection{Organization}
\label{Sec:sec:org}}

This review paper contains six sections.
Section ~\ref{Sec:over} gives an overview of GZSL, which includes the problem formulation, semantic information, embedding spaces and challenging issues.
Section~\ref{Sec:methods} {\color{black}reviews inductive and semantic transductive} GZSL methods, in which a hierarchical categorization of the GZSL methods is provided.
Each category is further divided into several constituents.
{\color{black} Section~\ref{Sec:sec:trans} focuses on the transductive GZSL methods.
Section~\ref{Sec:app} presents the applications of GZSL to various domains, including computer vision and natural language processing (NLP).}
A discussion on the research gaps and trends for future research, along with concluding remarks, is presented in Section~\ref{Sec:dis}.

\section{Overview of Generalized Zero-Shot Learning}
\label{Sec:over}
\begin{figure}
    \includegraphics[width=0.49\textwidth]{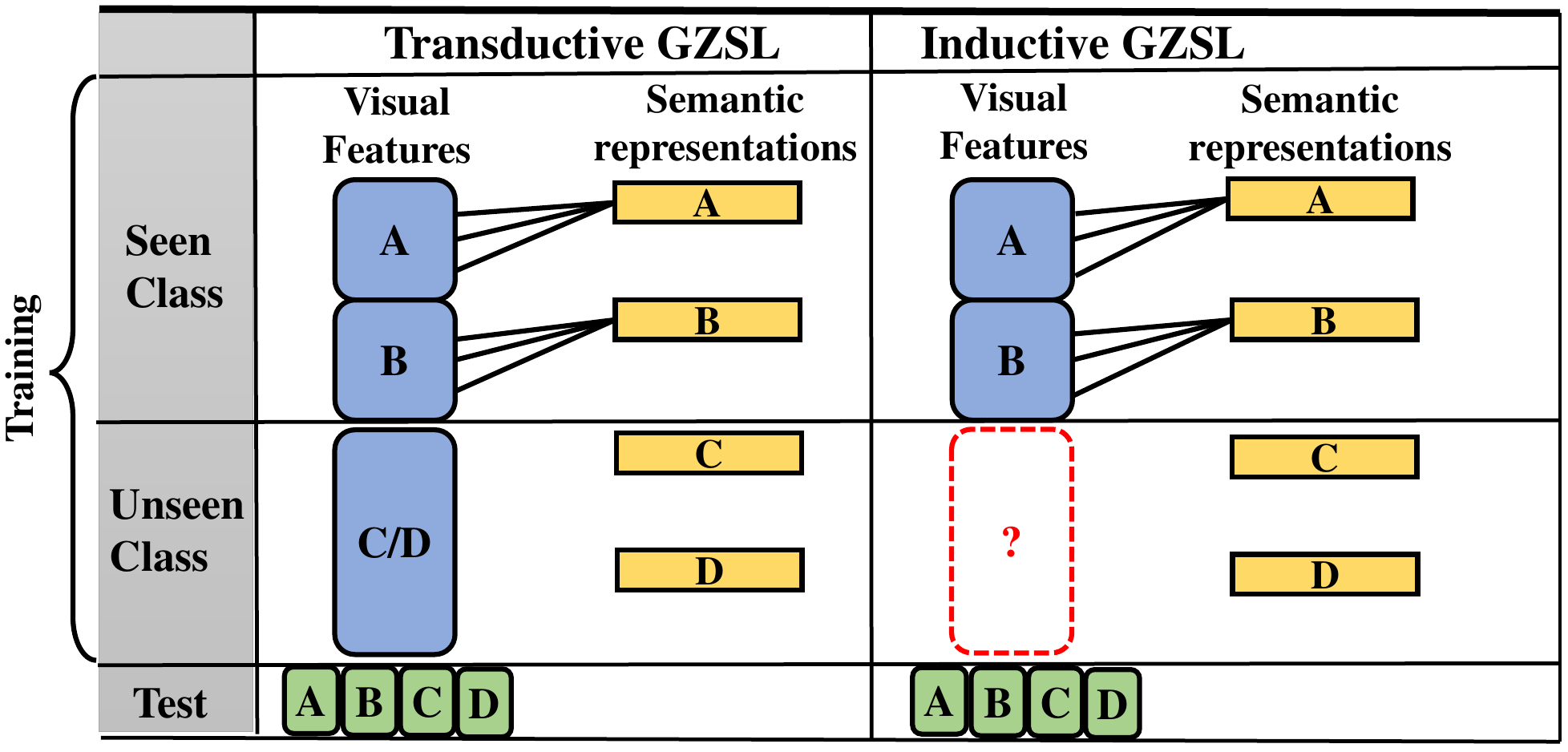}
    \caption{\color{black}A schematic view of Transductive and Inductive settings. In inductive setting, only the visual features and semantic representations of the seen classes ($A$ and $B$) are available. While transductive setting, in addition to the seen class information,  has access to the unlabelled visual samples of the unseen classes.}
    \label{fig:TZSLvsGZSL}
\end{figure}

\subsection{Problem Formulation}
\label{Sec:sec:problem}
Let $S=\{(x_i^s,a_i^s,y_i^s)_{i=1}^{N_s} | x_i^s\in X^s, a_i^s \in A^s, y_i^s \in Y^s\}$ and $U=\{(x_j^u,a_j^u,y_j^u)_{j=1}^{N_u} | x_j^s\in X^s, a_j^s \in A^s, y_j^s \in Y^s\}$ represent the seen and unseen class data sets, respectively, where $x_i^s, x_j^u \in \mathbb{R}^{D}$ indicate the $D$-dimensional images (visual features) in the feature space $\mathcal{X}$ that can be obtained using a pre-trained deep learning model such as ResNet~\cite{he2016deep}, VGG-19~\cite{simonyan2014very}, GoogLeNet~\cite{szegedy2015going};
$a_i^s,a_j^u \in \mathbb{R}^{K}$ indicate the $K$-dimensional semantic representations (i.e., attributes or word vectors) in the semantic space $\mathcal{A}$; $Y^s=\{y^s_1, ..., y^s_{C_s}\}$ and $Y^u=\{y^u_1, ..., y^u_{C_u}\}$ indicate the label sets of both seen and unseen classes in the label space $\mathcal{Y}$, where $C_s$ and $C_u$ are the number of seen and unseen classes, $\mathcal{Y}=Y^s \cup Y^u$ denotes the union of both seen and unseen classes and $Y^s\cap Y^u=\emptyset$. In GZSL, the objective is to learn a model $f_{GZSL}: \mathcal{X} \to \mathcal{Y}$ for classifying $N_t$ test samples, i.e., $D_{ts}=\{x_m,y_m\}_{m=1}^{N_t}$ where $x_m \in \mathbb{R}^{D}$, and $y_m \in \mathcal{Y}$.\par

The training phase of GZSL methods can be divided into two broad settings: inductive learning and transductive learning~\cite{xian2018zero}. Inductive learning utilises only the visual features and semantic information of the seen classes to build a model. In transductive setting, in addition to the seen class information, the semantic representations and unlabeled visual features of the unseen classes are exploited for learning~\cite{xian2018zero,zhu2019generalized}. Fig.~\ref{fig:TZSLvsGZSL} illustrates the main difference of transductive GZSL and inductive GZSL settings~\cite{zhang2018triple,yang2019self}. As can be seen, the prior of seen classes is available for both settings, but availability of the prior of unseen classes depends on the setting, as follows. In inductive learning setting, there is no prior knowledge of the unseen classes.  In transductive learning setting where the models use the unlabelled samples of the unseen classes, the prior of unseen classes is available. Although several frameworks have been developed under transductive learning~\cite{paul2019semantically,cheraghian2019transductive,rahman2019transductive,guan2018extreme,huo2018zero,long2018pseudo,zhang2020towards,xu2021holistically}, this learning paradigm is impractical.  On one hand, it violates the unseen assumption and reduces challenge.  On the other hand, it is not practical to assume that unlabeled data for all unseen classes are available. In addition, some studies~\cite{fu2015transductive,rahman2019transductive,cheraghian2019transductive,bo2021hardness} under transductive learning employ all image samples of the unseen classes during training, while others~\cite{song2018transductive,zhang2020deep} split the samples into two equal portions, one for training and another for inference.\par

Recently, several researchers~\cite{elhoseiny2019creativity,jha2021imaginative,yang2019self} argued that most of the generative-based methods, which are reviewed in Section~\ref{Sec:sec:gen}, are not pure inductive learning as the semantic information of the unseen classes is used to generate the visual features for the unseen classes. They categorized such generative-based methods as semantic transductive learning. In addition, they proposed inductive generative-based methods~\cite{elhoseiny2019creativity,jha2021imaginative,yang2019self} without accessing the semantic information of unseen classes before testing.

\subsection{Performance Indicators}
\label{Sec:sec:PI}
{\color{black}
To evaluate the performance of the GZSL methods, several indicators have been used in the literature.
Accuracy of seen ($Acc_s$) and accuracy of unseen ($Acc_u$) classes are two common performance indicators.
Chao et al.~\cite{chao2016empirical} introduced the area under seen-unseen accuracy curve (AUSUC) to balance the trade-off recognizing between seen and unseen classes for calibration-based techniques.
This curve can be obtained by varying $\gamma$ in equation (1) in the main text.
Techniques with higher AUSUC values aim to achieve a balanced performance in GZSL tasks.\par

Harmonic mean (HM) is another performance indicator that is able to measure the inherent biasness of GZSL-based methods with respect to the seen classes:
\begin{align} \label{eq:HM}
H=2*\frac{Acc_{s}*Acc_{u}}{Acc_{s}+Acc_{u}}.
\end{align}
If a GZSL method is biased towards the seen classes, its $Acc_s$ is higher than $Acc_u$ consequently the HM score drops down~\cite{rahman2018aunified}.}\par

\vspace{-0.42cm}
\subsection{Semantic Information}
\label{Sec:sem}
{\color{black} Semantic information is the key to GZSL. Since there are no labelled samples from the unseen classes, semantic information is used to build a relationship between both seen and unseen classes, thus making it possible to perform generalized zero-shot recognition. The semantic information must contain recognition properties of all unseen classes to guarantee that enough semantic information is provided for each unseen class. It also should be related to the samples in the feature space to guarantee usability of semantic information.
The idea of using semantic information is inspired by human recognition capability. Human can recognize samples from the unseen classes with the help of semantic information. As an example, a child can easily recognize zebra, if he/she has seen horses previously, and have the knowledge that a zebra looks like a horse with black and white strips.  Semantic information builds a space that includes both seen and unseen classes, which can be used to perform ZSL and GZSL. The most widely used semantic information for GZSL can be grouped into manually defined attributes~\cite{lampert2014attribute}, word vectors~\cite{mikolov2013distributed}, or their combinations.\par
}

\subsubsection{Manually defined attributes} These attributes describe the high-level characteristics of a class (category), such as shape (i.e., circle) and color (i.e., blue), which enable the GZSL model to recognize classes in the world.
The attributes are accurate, but require human efforts in annotation, which are not suitable for large-scale problems~\cite{kodirov2017semantic}.
Wu et al.~\cite{wu2019simple} proposed a global semantic consistency network (GSC-Net) to exploit the semantic attributes for both seen and unseen classes.
Lou et al.~\cite{Luo2020anovel} developed a data-specific feature extractor according to the attribute label tree.

\subsubsection{Word vectors} These vectors are automatically extracted from large text corpus (such as Wikipedia) to represent the similarities and differences between various words and describe the properties of each object.
Word vectors require less human labor, therefore they are suitable for large-scale data sets.
However, they contain noise which compromises the model performance.
As an example, Wang et al.~\cite{wang2019inductive} applied Node2Vec to produce the conceptualized word vectors.
The studies in~\cite{elhoseiny2017link,zhu2018agenerative,le2020webly,paz2020zest} attempted to extract semantic representations from noisy text descriptions, {\color{black} and Akata et al.~\cite{akata2016multi} proposed to extract semantic representations from multiple textual sources.}\par

{\color{black}
\vspace{-0.2cm}

\subsection{Embedding Spaces}
\label{Sec:sec:embd}
Most GZSL methods learn an embedding/mapping function to associate the low-level visual features of the seen classes with their corresponding semantic vectors.
This function can be optimized either via a ridge regression loss~\cite{yutaro2015ridge,daghaghi2019semantic} or ranking loss with respect to the compatibility scores of two spaces~\cite{rahman2018aunified}.
Then, the learned function is used to recognize novel classes by measuring the similarity level between the prototype representations and predicted representations of the data samples in the embedding space.
As every entry of the attribute vector represents a description of the class, it is expected that the classes with similar descriptions contain a similar attribute vector in the semantic space.
However, in the visual space, the classes with similar attributes may have large variations.
Therefore, finding such an embedding space is a challenging task, causing visual semantic ambiguity problems.\par

On the one hand, the embedding space can be divided into either Euclidean or non-Euclidean spaces.
While the Euclidean space is simpler, it is subject to information loss.
The non-Euclidean space, which is commonly based on graph networks, manifold learning, or clusters, usually uses the geometrical relation between spaces to preserve the relationships among the data samples~\cite{rezaei2020zero}.
On the other hand, the embedding space can be categorized into: \textit{semantic embedding}, \textit{visual embedding} and \textit{latent space embedding}~\cite{liu2020label}.
Each of these categories are discussed in the following subsections.\par

\begin{figure}
    \includegraphics[width=0.47\textwidth]{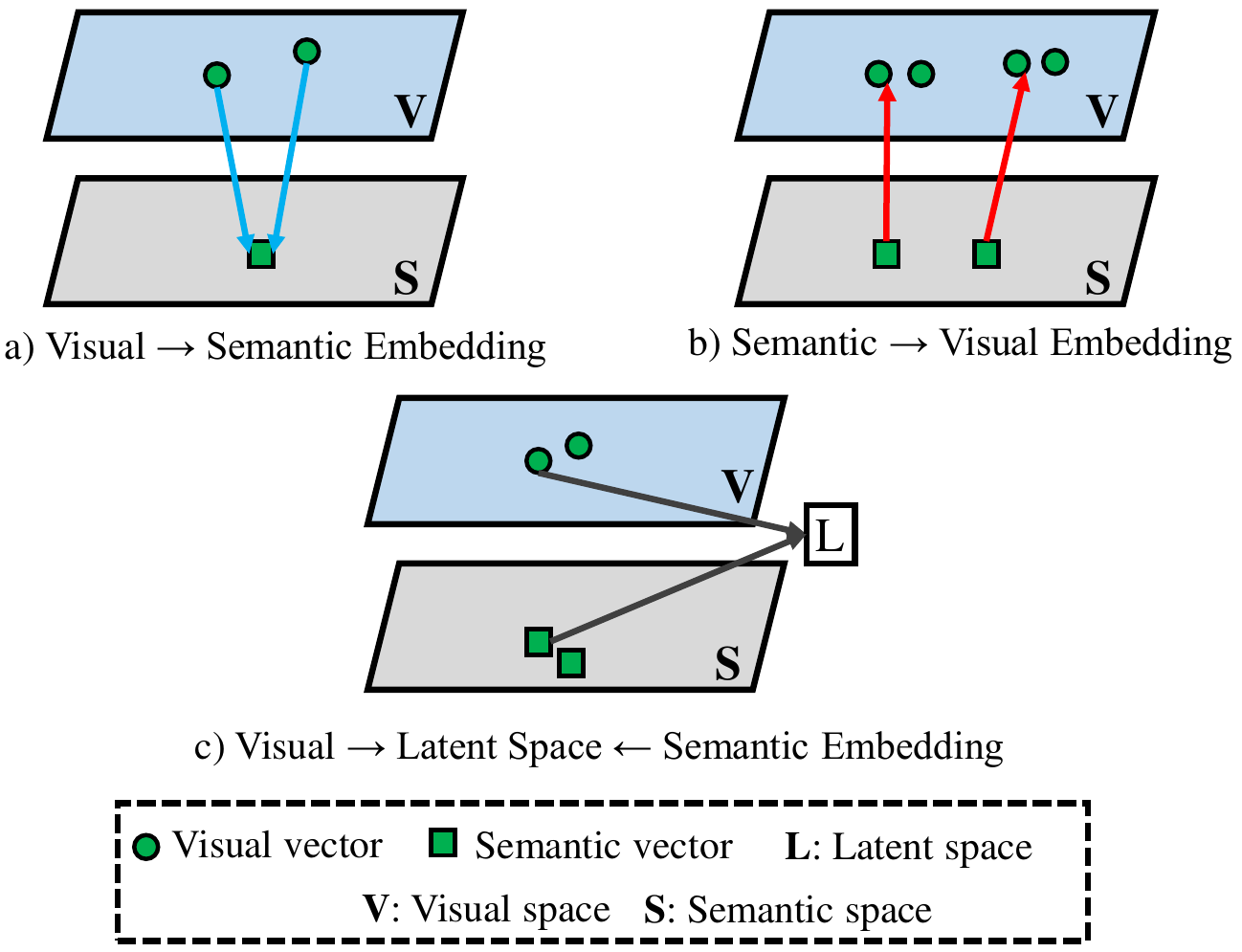}
    \caption{A schematic view of different embedding spaces in GZSL (adapted from~\cite{liu2020label}).}
    \label{fig:embd}
\end{figure}

\vspace{-0.2cm}
\subsubsection{Semantic Embedding}
\label{Sec:sec:sec:sem}
Semantic embedding (Fig~\ref{fig:embd} ({\color{green}a})) learns a (forward) projection function from the visual space to the semantic space using different constraints or loss functions, and perform classification in the semantic space.
The aim is to force semantic embedding of all images belonging to a class to be mapped to some ground-truth label embedding~\cite{chen2018zero,rahman2018aunified}.
Once the best projection function is obtained, the nearest neighbor search can be performed for recognition of a given test image.

\subsubsection{Visual Embedding}
\label{Sec:sec:sec:emd}
Visual embedding (Fig.~\ref{fig:embd} ({\color{green}b}))  learns a (reverse) projection function to map the semantic representations (back) into the visual space, and perform classification in the visual space. The goal is to make the semantic representations close to their corresponding visual features~\cite{yutaro2015ridge}.
After obtaining the best projection function, the nearest neighbor search can be used to recognize a given test image.


\subsubsection{Latent Embedding}
\label{Sec:sec:sec:latent}
Both semantic and visual embedding models learn a projection/embedding function from the space of one modality, i.e., visual or semantic, to the space of other modality.
However, it is a challenging issue to learn an explicit projection function between two spaces due to the distinctive properties of different modalities.
In this respect, latent space embedding (Fig.~\ref{fig:embd} ({\color{green}c})) projects both visual features and semantic representations into a common space $L$, i.e., a latent space, to explore some common semantic properties across different modalities~\cite{zhang2020towards,zhao2017zero,Yang2018dissimilarity}.
The aim is to project visual and semantic features of each class nearby into the latent space.
An ideal latent space should fulfill two conditions: \textit{(i)} intra-class compactness, and \textit{(ii)} inter-class separability~\cite{zhang2020towards}.
Introduced by Zhang et al.~\cite{zhang2017learning}, this mapping aims to overcome the hubness problem of ZSL models, which is discussed in Sub-section~\ref{Sec:sec:chal}.
}

\begin{figure}
\centering
    \includegraphics[width=0.4\textwidth]{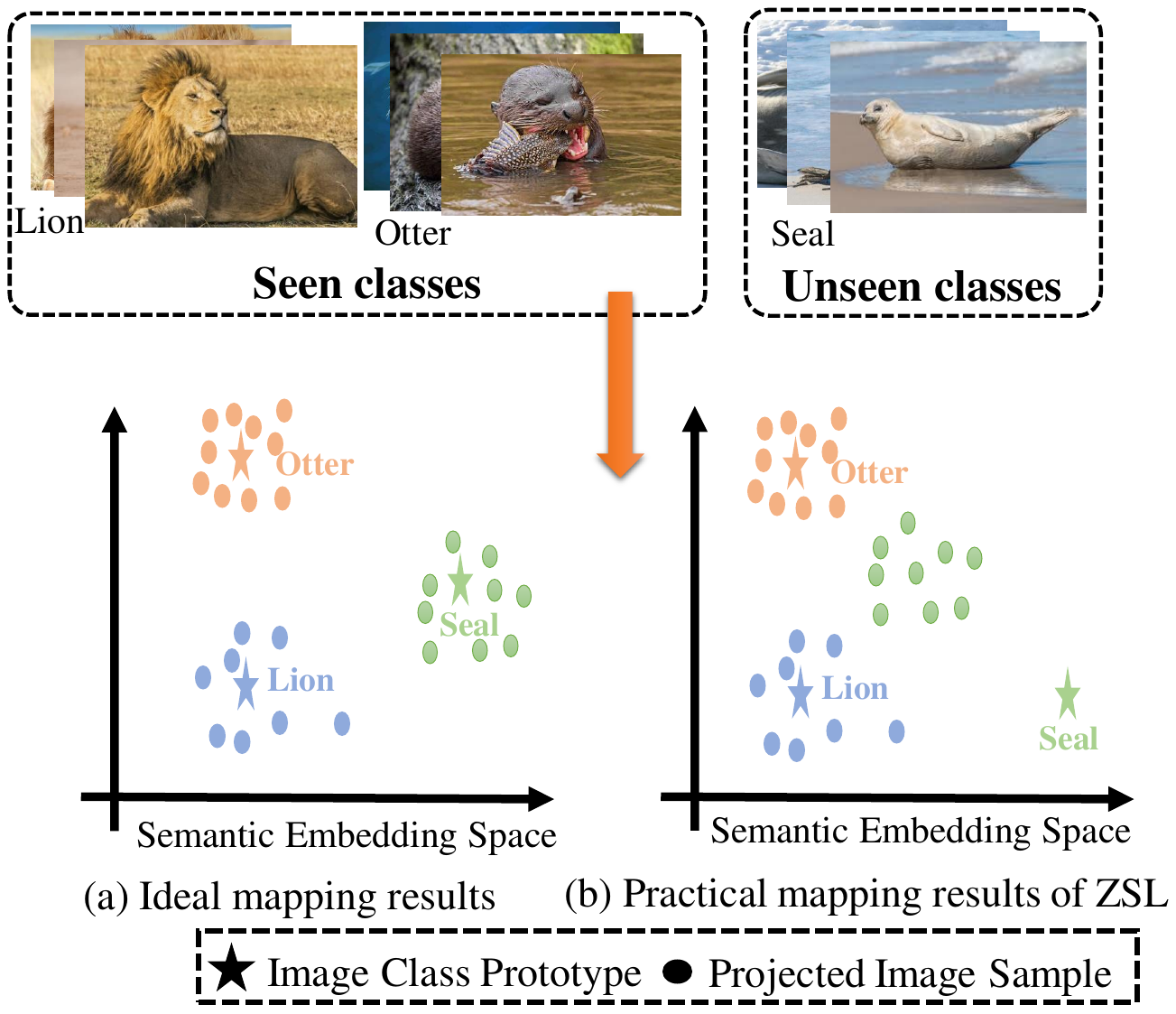}
    \caption{A schematic view of the projection domain shift problem (adapted from~\cite{jia2020deep}).}
    \label{fig:shift}
\end{figure}

\vspace{-0.2cm}
\subsection{Challenging Issues}
\label{Sec:sec:chal}
In GZSL, several challenging issues must be addressed.
The \textit{hubness problem}~\cite{kumar2018generalized,zhang2017learning} is one of the challenging issues of early ZSL and GZSL methods {\color{black} that learn semantic embedding space and utilize the nearest neighbor search to perform recognition.}
Hubness is an aspect of the curse of dimensionality that affects the nearest neighbors method, i.e., the number of times that a sample appears within the $k$-nearest neighbors of other samples ~\cite{radovanovic2010hubs}.
Dinu et al.~\cite{dinu2014improving} observed that a large number of different map vectors are surrounded by many common items, in which the presence of such items causes problem in high-dimensional spaces.\par

{\color{black}The \textit{projection domain shift problem} is another challenging issue of ZSL and GZSL methods. Both ZSL and GZSL models first leverage data samples of the seen classes to learn a mapping function between the visual and semantic spaces. Then, the learned mapping function is exploited to project the unseen class images from the visual space to semantic space. On the one hand, both visual and semantic spaces are two different entities. On the other hand, data samples of the seen and unseen classes are disjoint, unrelated for some classes, and their distributions can be different, resulting in a large domain gap.  As such, learning an embedding space using data samples from the seen classes without any adaptation to the unseen classes causes the projection domain shift problem~\cite{jia2020deep,fu2015transductive,zhao2017zero}. This problem is more challenging in GZSL, due to the existence of the seen classes during prediction. Since GZSL methods are required to recognize both seen and unseen classes during inference, they are usually biased towards the seen classes (which is discussed as the third problem of GZSL task in the next paragraph), because of the availability of visual features of seen classes during learning. Therefore, it is critical to learn a precise mapping function to avoid bias and ensure the effectiveness of the resulting GZSL models.
The \textit{projection domain shift} problem differs from the \textit{vanilla domain shift} problem. Unlike the vanilla (conventional) domain shift problem where two domains share the same categories~\cite{liang2021domain} and only the joint distribution of input and output differs between the training and test stages~\cite{quinonero2008dataset}, the projection domain shift can be directly observed in terms of the projection shift, rather than the feature distribution shift. In addition, the seen (source) domain classes are completely different from the unseen (target) domain classes, and both can even be unrelated~\cite{fu2015transductive}.}\par

{\color{black}Fig.~\ref{fig:shift} (a) shows an ideal unbiased mapping function that forces the projected visual samples of both seen and unseen classes to surround their own semantic features in the latent space. In practice, the training and test samples are disjoint in GZSL tasks.  This results in learning an unbiased mapping function for the seen classes, which can project the visual features of the unseen classes far away from their semantic features (see Fig.~\ref{fig:shift} (b)).}
This problem is more common in inductive-based methods, as they have no access to the unseen class data during training.
To overcome this problem, inductive-based methods incorporate additional constraints or information from the seen classes.
Besides that, several transdactive-based methods have been developed to alleviate the {\color{black}projection} domain shift problem~\cite{cheraghian2019transductive,rahman2019transductive,guan2018extreme,huo2018zero}.
These methods use the manifold information of the unseen classes for learning.\par

\begin{figure}
    \includegraphics[width=0.5\textwidth]{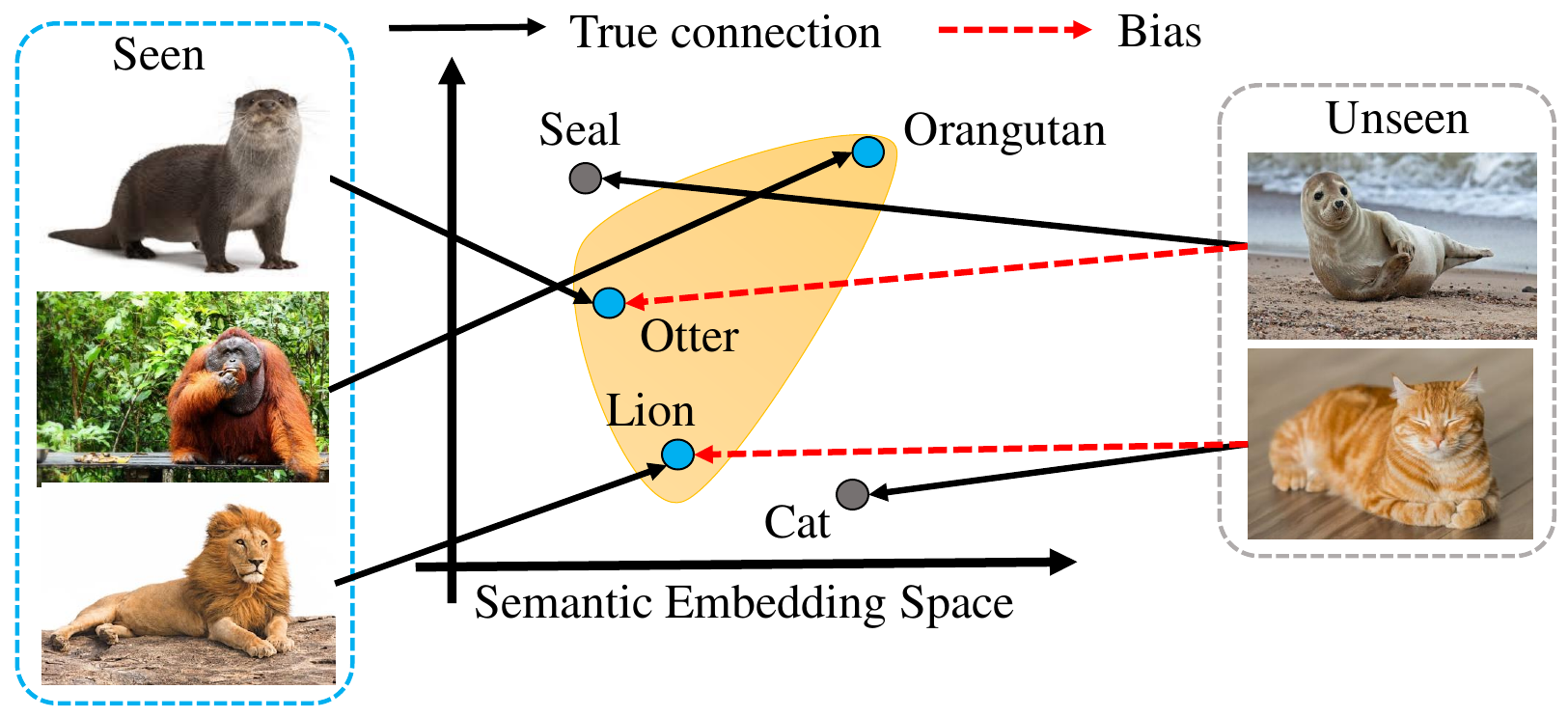}
    \caption{A schematic view of the bias concerning seen classes (source) in the semantic embedding space (adapted from~\cite{zhang2019co}).}
    \label{fig:ses}
\end{figure}

Detectors aim to identify whether a test sample belongs to the seen or unseen classes.
This strategy limits the set of possible classes by providing information to which set (seen or unseen) a test sample belongs to.
Socher et al.~\cite{socher2013zero} considered that the unseen classes are projected to out-of-distribution (OOD) with respect to the seen ones.
Then, data samples from unseen classes are treated as outliers with respect to the distribution of the seen classes.
Bhattaxharjee~\cite{bhattacharjee2019autoencoder} developed an auto-encoder-based framework to identify the set of possible classes.
To achieve this, additional information, i.e., the correct class information, is imposed into the decoder to reconstruct the input samples.\par

Since GZSL methods use the data samples from the seen classes to learn a model to perform recognition for both seen and unseen classes, they are usually \textit{biased towards the seen classes}, leading to misclassification of data from the unseen classes into the seen classes (see Fig.~\ref{fig:ses}), in which most of the ZSL methods cannot effectively solve this problem~\cite{zhang2019co}.
To mitigate this issue, several strategies have been proposed, such as calibrated stacking~\cite{chao2016empirical,felix2019multi} and novelty detector~\cite{socher2013zero,bhattacharjee2019autoencoder,atzmon2019adaptive,min2020domain}.
The \textit{calibrated stacking}~\cite{chao2016empirical} method balances the trade-off between recognizing data samples from both seen and unseen classes using the following formulation.
\begin{align} \label{eq:CalStack}
\hat{y}=\arg \max_{c\in \mathcal{Y}} f_c(x)-\gamma \amalg[c\in Y^s],
\end{align}
where $\gamma$ is a calibration factor and $\amalg[\cdot]\in \{0,1\}$ indicates whether $c$ is from seen classes or otherwise.
In fact, $\gamma$ can be interpreted as the prior likelihood of a sample from the unseen classes.
When $\gamma \to -\infty$, the classifier will classify all data samples into one of the seen classes, and vice versa.
Le et al.~\cite{le2019classical} proposed to find an optimal $\gamma$ that balances the trade-off between accuracy of the seen and unseen classes.
Later, several studies used the calibrated stacking technique to solve the GZSL problem~\cite{felix2019multi,changpinyo2020classifier,xu2020attribute,guo2019dual,niu2018zero}.
Similar to the calibrated stacking, \textit{scaled calibration}~\cite{das2019zero} and \textit{probabilistic representation}~\cite{huynh2020fine,oreshkin2020clarel} have been proposed to balance the trade-off between both seen and unseen classes.
Studies~\cite{liu2018generalized,felix2020augmentation} made the unseen classes more confident and the seen classes less confident using temperature scaling~\cite{hinton2015distilling}.\par

Detectors aim to identify whether a test sample belongs to the seen or unseen classes.
This strategy limits the set of possible classes by providing information to which set (seen or unseen) a test sample belongs to.
Socher et al.~\cite{socher2013zero} considered that the unseen classes are projected to out-of-distribution (OOD) with respect to the seen ones.
Then, data samples from unseen classes are treated as outliers with respect to the distribution of the seen classes.
Bhattaxharjee~\cite{bhattacharjee2019autoencoder} developed an auto-encoder-based framework to identify the set of possible classes.
To achieve this, additional information, i.e., the correct class information, is imposed into the decoder to reconstruct the input samples.\par

Later, entropy-based~\cite{min2020domain}, probabilistic-based~\cite{wang2020domain,atzmon2019adaptive}, distance-based~\cite{li2020improving}, cluster-based~\cite{hayashi2021cluster} and parametric novelty detection~\cite{wu2019simple} approaches have been developed to detect OOD, i.e., the unseen classes.
Felix et al.~\cite{felix2019generalised} learned a discriminative model using the latent space to identify whether a test sample belongs to a seen or unseen class.
Geng et al.~\cite{geng2019visual} decomposed GZSL into \textit{open set recognition} (OSR)~\cite{scheirer2013toward} and ZSL tasks.

\vspace{-0.32cm}
\section{Review of GZSL Methods}
\label{Sec:methods}

The main idea of GZSL is to classify objects of both seen and unseen classes by transferring knowledge from the seen classes to the unseen ones through semantic representations. To achieve this, two key issues must be addressed: \textit{(i)} how to transfer knowledge from the seen classes to unseen ones; \textit(ii) how to learn a model to recognize images from both seen and unseen classes without having access to the labeled samples of unseen classes~\cite{liu2018generalized}.
In this regard, many methods have been proposed, which can be broadly categorized into:

\begin{figure}
\centering
        \begin{subfigure}[b]{0.46\textwidth}
            \centering
            \includegraphics[width=\textwidth]{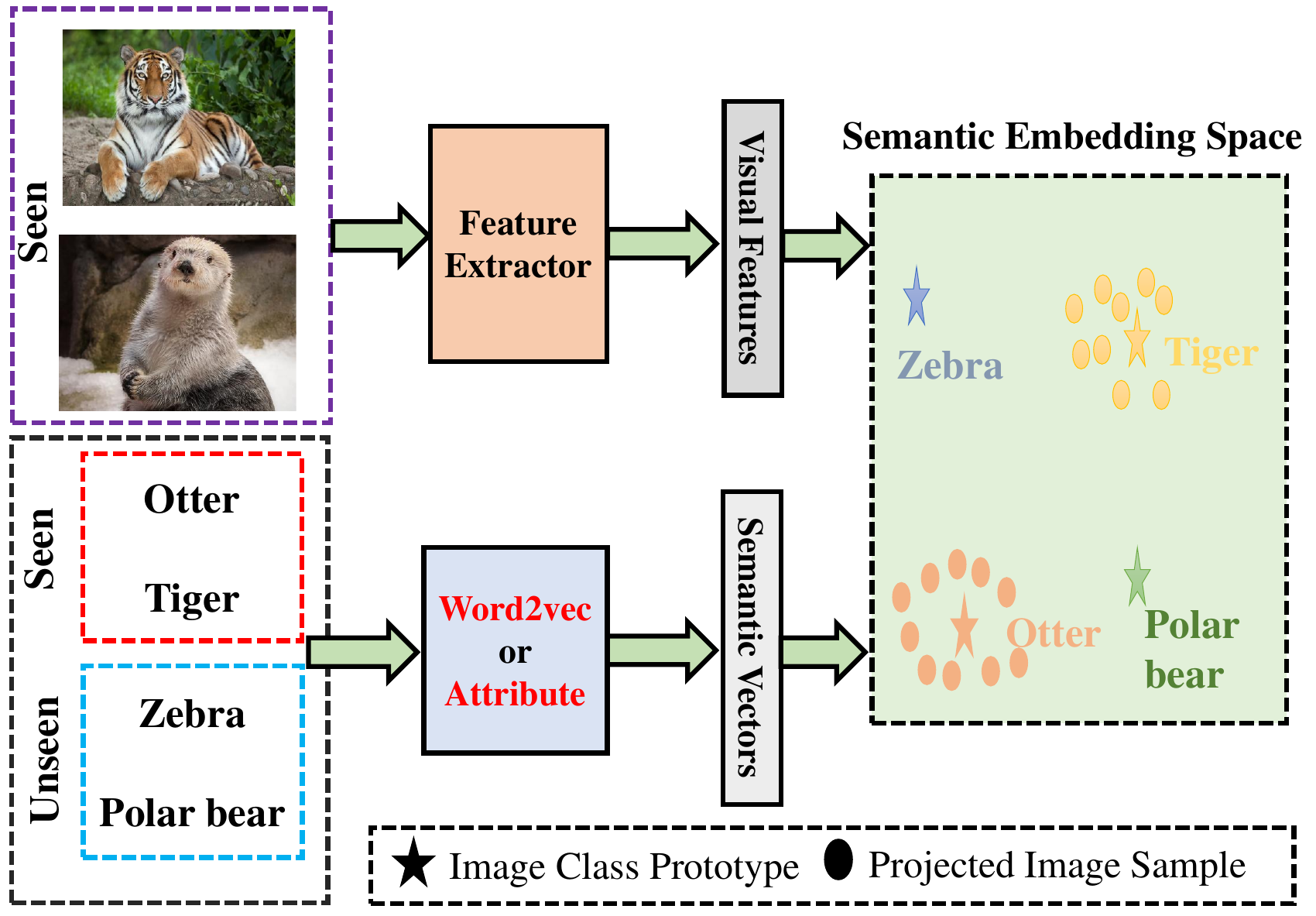}
            \caption{}
            \label{fig:emd}
    \end{subfigure}

    \vspace*{0.5cm}
    \begin{subfigure}[b]{0.46\textwidth}
            \centering
            \includegraphics[width=\textwidth]{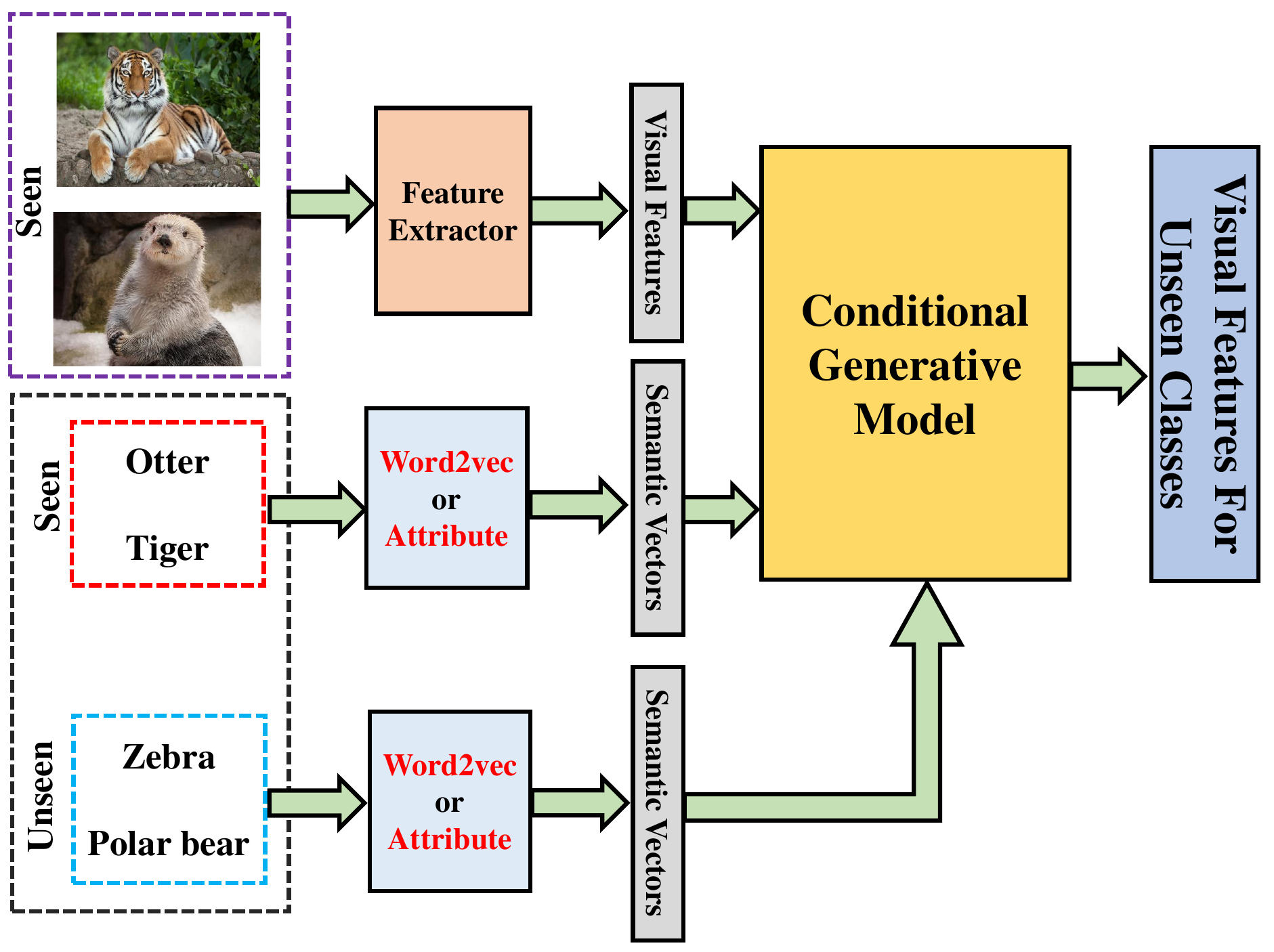}
            \caption{}
            \label{fig:gen}
    \end{subfigure}
    \vspace*{0.5cm}
    \caption{\color{black}Embedding-based versus generative-based methods. The embedding based methods (a) lean an embedding space to project the visual and semantic features of seen classes into a common space.
    Then, the learned embedding space is used to perform recognition.
    In contrast, the generative-based methods (b) learn a generative model based on samples of seen classes conditioned on their semantic features.
    Then, the learned model is used to generate visual features for unseen classes using the semantic features of unseen classes.}\label{fig:EmVsGen}
\end{figure}

\begin{figure*}[hbt!]
 \begin{center}
 \includegraphics[width=0.99\textwidth]{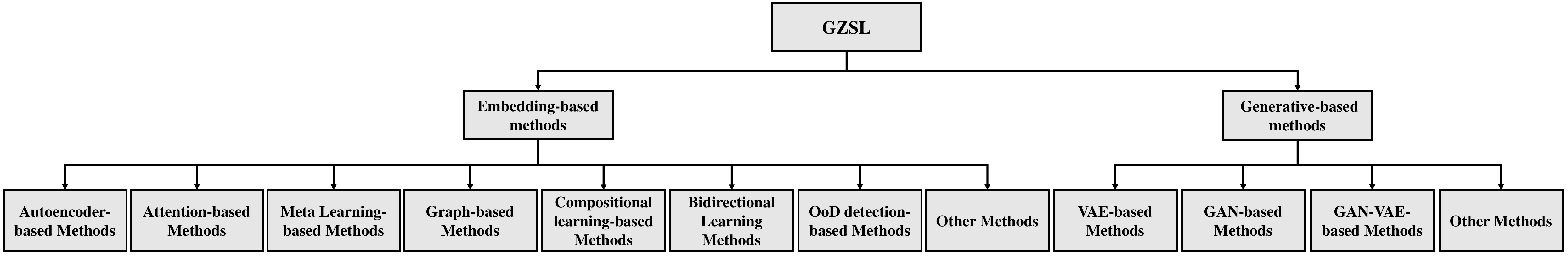}
  \end{center}
  \caption{The taxonomy of GZSL models.}
  \label{Fig:taxo}
\end{figure*}

\begin{itemize}
\item\textbf{Embedding-based methods:} learn an embedding space to associate the low-level visual features of seen classes with their corresponding semantic vectors.
The learned projection function is used to recognize novel classes by measuring the similarity level between the prototype representations and predicted representations of the data samples in the embedding space (see Fig.~\ref{fig:EmVsGen} ({\color{green}a})).

\item\textbf{Generative-based methods:} learn a model to generate images or visual features for the unseen classes based on the samples of seen classes and semantic representations of both classes.
By generating samples for unseen classes, a GZSL problem can be converted into a conventional supervised learning problem (see Fig.~\ref{fig:EmVsGen} ({\color{green}b})).
Based on single homogenous process, a model can be trained to classify the test samples belonging to both seen and unseen classes and solve the bias problem.
Although these methods perform recognition in the visual space and they can be categorized as visual embedding models, we separate them from the embedding based methods.
\end{itemize}

A hierarchical categorization of both methods, together with their sub-categories, is provided in Fig.~\ref{Fig:taxo}.\par

\vspace{-0.2cm}
\subsection{Embedding-based Methods}
\label{Sec:embd}
In recent years, various embedding-based methods have been used to formulate a framework to tackle GZSL problems.
These methods can be divided into graph-based, attention-based, autoencoder-based, meta learning,  compositional learning and bidirectional learning methods, as shown in Fig.~\ref{Fig:taxo}.
In the following subsections, we review each of these categories, and provide a summary of these methods in Table~\ref{Table:embeding}.\par

\vspace{-0.15cm}
{\color{black}
\subsubsection{Out-of-distribution detection-based methods}
\label{Sec:sec:OOD}
Out-of-distribution or outlier detection aims to identify data samples that are abnormal or significantly different from other available samples. Several studies applied outlier detection techniques to solve GZSL tasks~\cite{socher2013zero,lee2018hierarchical,atzmon2019adaptive,bhattacharjee2019autoencoder}. Firstly, outlier detection techniques are employed to separate the seen class instances from those of the unseen classes. Then, domain expert classifiers (seen/unseen), e.g., standard classifiers for seen classes and ZSL methods for unseen classes, are adopted to separately classify seen and unseen class data samples. Lee et al.~\cite{lee2018hierarchical} developed a novelty detector based on a hierarchical taxonomy based on language information. Their method builds a taxonomy with the hyper-nymhyponym relationships between known classes, in a way that objects of unseen classes are expected to be categorized into one of the most relevant seen classes. Atzmon and Chechik~\cite{atzmon2019adaptive} devised a probability based gating mechanism and introduced a Laplace-like prior into the gating mechanism to distinguish seen class samples from the unseen ones. However, training the gate is a challenging issue since visual features from the unseen classes are unavailable during training. To solve this issue, boundary-based OOD classifier~\cite{chen2020boundary}, semantic encoding classifier~\cite{ding2021semantic} and domain detector based on entropy gate~\cite{bhattacharjee2019autoencoder} have been proposed to separate seen class samples from the unseen ones. In addition, the studies in~\cite{mandal2019out,keshari2020generalized} integrated OOD techniques into generative methods to tackle GZSL tasks.
}

\vspace{-0.15cm}
\subsubsection{Graph-based Methods}
\label{Sec:sec:graph}
\begin{figure*}
\centering
    \includegraphics[width=0.86\textwidth]{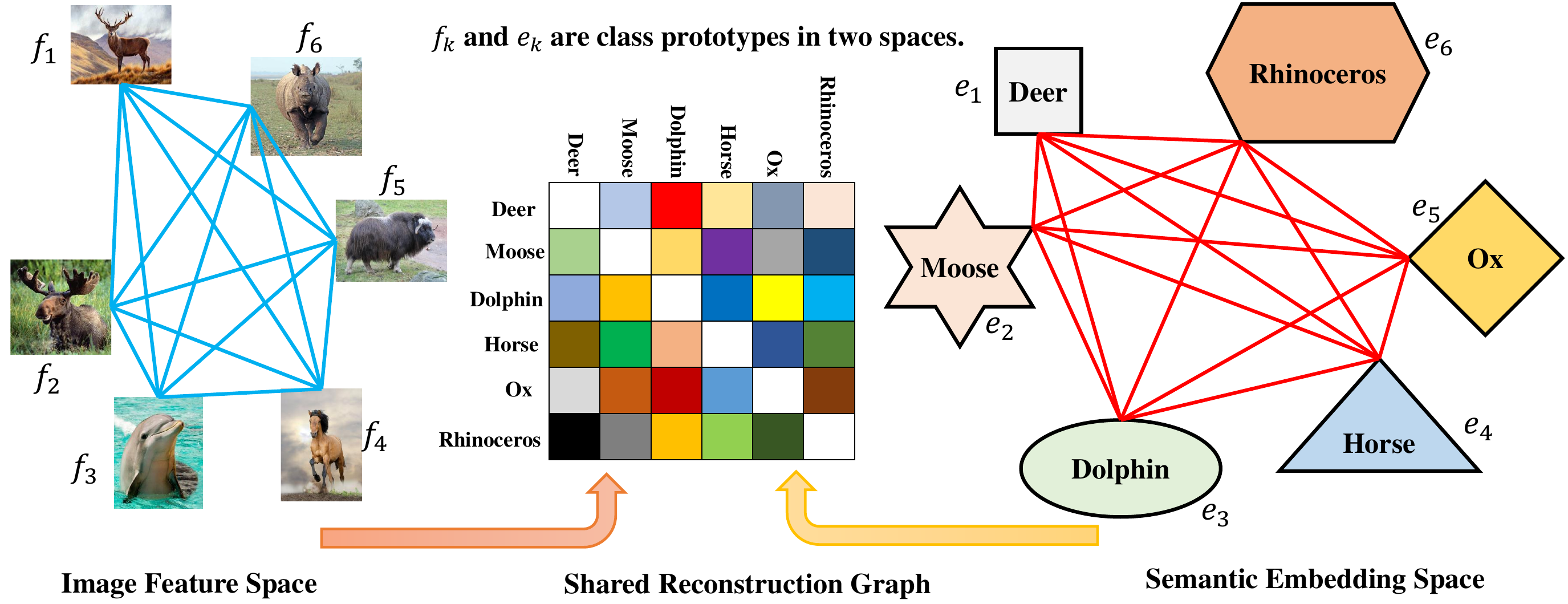}
    \caption{\color{black} A schematic view of SRG~\cite{zhao2017zero}. Firstly, the image prototype $f_k$ and semantic prototype $e_k$ of each class are reconstructed using the cluster center and (\ref{eq:SRG}), respectively. Then, the shared reconstruction coefficients between two spaces is learned. Finally, the learned SRG is used to synthesize class prototypes for unseen classes to perform prediction. }
    \label{fig:SRG}
\end{figure*}

{\color{black}
Graphs are useful for modelling a set of objects with a data structure consisting of nodes and their relationships (edges)~\cite{zhou2020graph}. Graph learning leverages machine learning techniques to extract relevant features, mapping properties of a graph into a feature vector with the same dimensions in the embedding space. Machine learning techniques convert graph-based properties into a set of features without projecting the extracted information into a lower dimensional space~\cite{xia2021graph}.  Generally, each class is represented as a node in a graph-based method. Each node is connected to other nodes (i.e., classes) through edges that encode their relationships.  The geometric structure of features in the latent space is preserved in a graph, leading to a compact representation of richer information, as compared with other techniques~\cite{zhao2017zero}. Nonetheless, learning a classifier using structured information and complex relationships without visual examples for the unseen classes is a challenging issue, and the use of graph-based information increases the model complexity.


Recently, graph learning techniques have shown effective paradigm in GZSL~\cite{zhao2017zero,wang2019asymmetric,xu2021semi,bhagat2021novel,samplawski2020zero,li2020transferrable}.}
For example, the \textit{shared reconstructed graph} (SRG)~\cite{zhao2017zero} (Fig.~\ref{fig:SRG}) uses the cluster center of each class to represent the class prototype in the image feature space, and reconstructs each semantic prototype as follows:
\begin{align} \label{eq:SRG}
    e_k=\sum_{i=1}^{K}a_{k}^{i}b_{k}^{i}=Ab_k,~s.t.~b_k^k=0,
\end{align}
where $b_k \in \mathbb{R}^{K\times 1}$ includes the reconstruction coefficients, and $K=1,...,C_s+C_u$.
After learning the relationships among classes, the shared reconstruction coefficients between two spaces is learned to synthesize image prototypes for the unseen classes.
SRG contains a sparsity constraint, which enables the model to divide the classes into many clusters of different subspaces.A regularization term is adopted to select fewer and relevant classes during the reconstruction process.
The reconstruction coefficients are shared, in order to transfer knowledge from the semantic prototypes to image prototypes. In addition, the unseen semantic embedding method is used to mitigate the projection domain shift problem, in which the graph of the seen image prototypes is adopted to alleviate the space shift problem.\par
AGZSL~\cite{wang2019asymmetric}, i.e., \textit{asymmetric graph-based ZSL}, combines the class-level semantic manifold with the instance-level visual manifold by constructing an asymmetric graph.
In addition, a constraint is made to project the visual and attribute features orthogonally when they belong to different classes.
The studies in~\cite{wang2018zero,wang2019inductive,li2020anchor,xie2020region} exploit the graph convolutional network (GCN)~\cite{kipf2016semi} to transfer knowledge among different categories.
In~\cite{wang2018zero,wang2019inductive}, GCN is applied to generate super-classes in the semantic space.
The cosine distance is used to minimize the distance between the visual features of the seen classes and corresponding semantic representations, and a triplet margin loss function is optimized to avoid the hubness problem.
\textit{Discriminative anchor generation and distribution alignment} (DAGDA)~\cite{li2020anchor} uses a diffusion-based GCN to generate anchors for each category.
Specifically, a \textit{semantic relation regularization} is derived to refine the distribution in the anchor space.
To mitigate the hubness problem, two auto-encoders are employed to keep the original information of both features in the latent space.
Besides that, Xie et al.~\cite{xie2020region} devised an attention technique to find the most important regions of the image and then used these regions as a node to represent the graph.\par

\vspace{-0.15cm}
\subsubsection{Meta Learning-based Methods}
\label{Sec:sec:meta}
{\color{black}Meta learning, which is also known as learning to learn, is a subset of learning paradigms that learns from other learning algorithms. It aims to extract transferable knowledge from a set of auxiliary tasks, in order to devise a model while avoiding the overfitting problem.  The underlying principle of a meta learning method helps identify the best learning algorithm for a specific data set. Meta learning improves the performance of learning algorithms by changing some aspects according to the experimental results and by optimizing the number of experiments.
}
Several studies~\cite{sung2018learning,hu2018correction,verma2021meta,verma2020meta,yu2020episode,liu2021task,verma2021towards} utilized meta learning strategy to solve GZSL problems.
Meta learning based GZSL methods divide the training classes into two sets, i.e., support and query, which correspond to the seen and unseen classes.
Different tasks are trained by randomly selecting the classes from both the support and query sets.
This mechanism helps meta learning methods to transfer knowledge from the seen to unseen classes, therefore alleviating the bias problem ~\cite{liu2021task}.\par

Sung et al.~\cite{sung2018learning} trained an auxiliary parameterized network using a meta learning method.
The aim is to paramaterize a feedforward neural network for GZSL.
Specifically, a relation module is devised to compute the similarity metric between the output of a cooperation module and the feature vector of a data sample.
Then, the learned function is used for recognition.
Introduced in~\cite{hu2018correction}, the meta-learning method consists of a task module to provide an initial prediction and a correction module to update the initial prediction.
Various task modules are designed to learn different subsets of training samples. Then, the prediction from a task module is updated through training of the correction module.
Other examples of generative-based meta-learning modules for GZSL are also available, e.g.~\cite{verma2020meta,yu2020episode,verma2021towards,liu2021task}.\par

\vspace{-0.1901cm}
\subsubsection{Attention-based Methods}
\label{Sec:sec:attention}
 \color{black}{Unlike other methods that learn an embedding space between global visual features and semantic vectors, attention-based methods focus on learning the most important image regions. In other words, the attention mechanism seeks to add weights into deep learning models as trainable parameters to augment the most important parts of the input, e.g., sentences and images.  In general, attention-based methods are effective for identifying fine-grained classes because these classes contain discriminative information only in a few regions.  Following the general principle of an attention-based method, an image is divided into many regions, and the most important regions are identified by applying an attention technique~\cite{huynh2020fine,xie2019attentive}. One of the major advantages of the attention mechanism is its capability to recognize important information pertinent to performing a task in an input, leading to improved results. On the other hand, the attention mechanism generally increases computational load, affecting real-time implementation of attention-based methods.}\par

\begin{figure*}
    \includegraphics[width=0.97\textwidth]{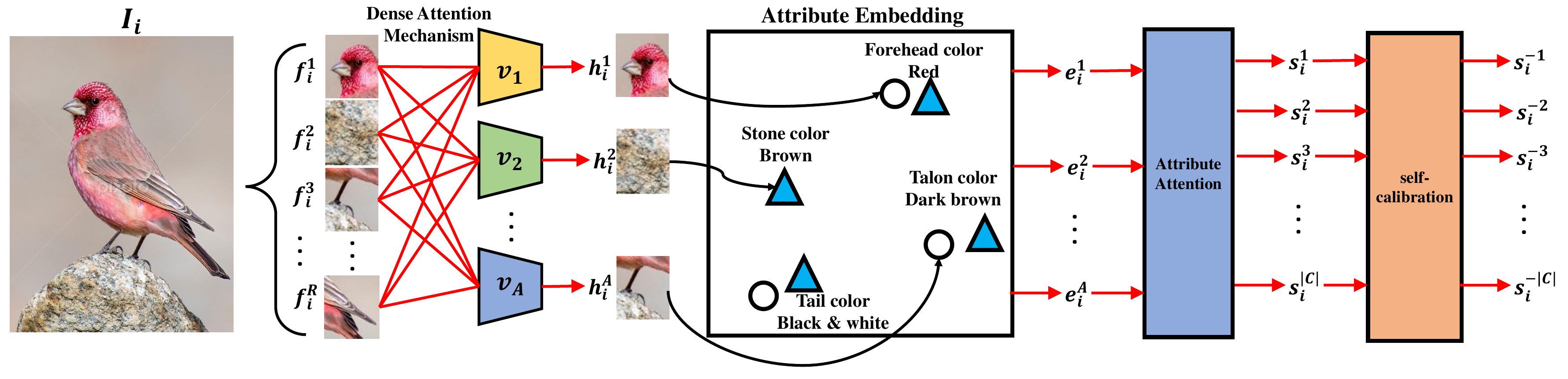}
    \caption{\color{black}A schematic of DAZLE~\cite{huynh2020fine}. After extracting image features of the $R$ regions, the attention features of all attributes are computed using a dense attention mechanism. Then, the attention features are aligned with the attribute semantic vectors, in order to compute the score of attributes in the image.}
    \label{fig:DAZLE}
\end{figure*}

The \textit{dense attention zero-shot learning} (DAZLE)~\cite{huynh2020fine} (Fig.~\ref{fig:DAZLE}) obtains visual features by focusing on the most relevant regions pertaining to each attribute.
Then, an \textit{attribute embedding technique} is devised to adjust each obtained attribute feature with its corresponding attribute semantic vector.
A similar approach is applied to solve multi-label GZSL problems in~\cite{huynh2020ashared}.
The \textit{attentive region embedding network} (AREN)~\cite{xie2019attentive} incorporates an \textit{attentive compressed second-order embedding} mechanism to automatically discover the discriminative regions and capture the second-ordered appearance differences.
In~\cite{ji2018stacked}, semantic representations are used to guide the visual features to generate an attention map.
In~\cite{yang2019self}, an embedding space was formulated by measuring the focus ratio vector for each dimension using a self-focus mechanism.
Zhu et al.~\cite{zhu2019generalized} used a multi-attention model to map the visual features into a finite set of feature vectors.
Each feature vector is modeled as a $C$-dimensional Gaussian mixture model with the isotropic components.
Using these low-dimensional embedding mechanisms allows the model to focus on the most important regions of the image as well as remove irrelevant features, therefore reducing the semantic-visual gap.
In addition, a visual oracle is proposed for GZSL to reduce noise and provide information on the presence/absence of classes.\par

The \textit{gaze estimation module} (GEM)~\cite{liu2021goal}, which is inspired by the gaze behavior of human, i.e., paying attention to the parts of an object with discriminative attributes, aims to improve the localization of discriminative attributes. In~\cite{yang2020simple}, the localized attributes are used for projecting the local features into the semantic space.  It exploits a global average pooling scheme as an aggregation mechanism to further reduce the bias (since the obtained patterns for the unseen classes are similar to those of seen ones) and improve localization. In contrast, the \textit{semantic-guided multi-attention} (SGMA) localization model~\cite{zhu2019semantic} jointly learns from both global and local features to provide a richer visual expression.\par

Liu et al.~\cite{liu2020attribute} used a \textit{graph attention network}~\cite{shen2017disan} to generate optimized attribute vector for each class.
Ji et al.~\cite{ji2019semantics} proposed a \textit{semantic attention network} to select the same number of visual samples from each training class to ensure that each class contributes equally during each training iteration.
In addition, the model proposed in~\cite{yeh2020zero} searches for a combination of semantic representations to separate one class from others.
On the other hand, Paz et al.~\cite{paz2020zest} designed \textit{visually relevant language}~\cite{winn2016detecting} to extract relevant sentences to the objects from noisy texts.\par

\vspace{-0.2cm}
\subsubsection{Compositional learning-based Methods}
\label{Sec:sec:comp}
Compositional learning (CL) aims to learn a model that can recognize a combinations of unseen compositions of known objects, e.g., fish and cat, and primitive states, e.g., cute and old~\cite{misra2017red,naeem2021learning}.
Recently, the concept of CL has been applied to ZSL, known as compositional ZSL.
Kato et al.~\cite{kato2018compositional} introduced a framework to recognize zero-shot human actions.
A GCN is used to build an external knowledge graph based on the extracted subject, verb and object (SVO) triplets~\cite{chen2013neil} from knowledge bases, in order to record a large range of human object interactions.
Each node in the graph indicates a noun (object) or a verb (motion) with
word embedding as its feature.
Each action node, which is represented by a SVO-triplet, propagates information along the graph to learn its representations.
Finally, both visual features and learned graph are jointly projected to a latent space for zero-shot recognition of human actions.\par

The \textit{task-driven modular network} (TMN)~\cite{purushwalkam2019task} employs a modular structure to transfer concepts in the high-level semantic spaces of CNNs and extracts features that are related to all members of the input triplet to determine the \textit{joint-compatibility} among visual features and object attributes.
Sylvain et al.~\cite{sylvain2019locality} empirically highlighted the importance of focusing on local image regions (via attention) as well as combining local knowledge to recognize the unseen classes.
Their work has been further supported by subsequent studies~\cite{advances2020huynh,atzom2020acausal}.\par

\vspace{-0.15cm}
\subsubsection{Bidirectional Learning Methods}
\label{Sec:sec:dual}
This category leverages the bidirectional projections to fully utilize information in data samples and learn more generalizable projections, in order to differentiate between the seen and unseen classes~\cite{guo2019dual,liu2020label,liu2019joint,ji2020dual}.
In~\cite{liu2019joint}, the visual and semantic spaces are jointly projected into a shared subspace, and then each space is reconstructed through a bidirectional projection learning.
The \textit{dual-triplet network} (DTNet)~\cite{ji2020dual} uses two triplet modules to construct a more discriminative metric space, i.e., one considers the attribute relationship between categories by learning a mapping from the attribute space to visual space, while another considers the visual features.\par


Guo et al.~\cite{guo2019dual} considered a dual-view ranking by introducing a loss function that jointly minimizes the image view labels and label-view image rankings.
This dual-view ranking scheme enables the model to train a better image-label matching model.
Specifically, the scheme ranks the correct label before any other labels for a training sample, while the label-view image ranking aims to rank the respective images to their correct classes before considering the images from other classes.
In addition, a density adaptive margin is used to set a margin based on the data samples.
This is because the density of images varies in different feature spaces, and different images can have different similarity scores.
The model proposed in~\cite{liu2020label} consists of two parts: \textit{(i)} \textit{visual-label activating} that learns a embedding space using regression; \textit{(ii)} \textit{semantic-label activating} that learns a projection function from the semantic space to the label space.
In addition, a bidirectional reconstruction constraint between the semantic and labels is added to alleviate the projection shift problem.\par

Zhang et al.~\cite{zhang2018triple} explained a class level overfitting problem.
It is related to parameter fitting during training without prior knowledge about the unseen classes.
To solve this problem, a \textit{triple verification network} (TVN) is used for addressing GZSL as a \textit{verification} task.
The verification procedure aims to predict whether a pair of given samples belongs to the same class or otherwise.
The TVN model projects the seen classes into an orthogonal space, in order to obtain a better performance and a faster convergence speed.
Then, a dual regression (DR) method is proposed to regress both visual features and semantic representations to be compatible with the unseen classes.\par

\subsubsection{Autoencoder-based Methods}
\label{Sec:sec:auto}
{\color{black}
Autoencoders (AEs) are unsupervised learning techniques that leverage NNs for representation learning. They learn how to compress/encode the data firstly. Then, they learn how to reconstruct the data back into a representation as close to the original data as possible. In other words, AEs exploit an encoder to learn an embedding space and then employ a decoder to reconstruct the inputs. The main advantage of AEs is that they can be trained in an unsupervised manner.

AEs have been widely used to solve GZSL problem. To achieve this, a decoder can be imposed by an additional constraint for learning different mappings~\cite{liu2018zero}.}
The framework proposed by Biswas and Annadani~\cite{biswas2018preserving} integrates the similarity level, e.g., cosine distance, into an objective function to preserve the relations.
\textit{Latent space encoding} (LSE)~\cite{yu2018zero} explores some common semantic characteristics between different modalities and connects them together.
For each modality, an encoder-decoder framework is exploited, in which an encoder is used to decompose the inputs as a latent representation while a decoder is employed to reconstruct the inputs.
\textit{Product quantization ZSL} (PQZSL)~\cite{li2019compressing} defines an orthogonal common space, which learns a codebook, to project the visual features and semantic representations into a common space using a center loss function~\cite{guo2017sinet} and an auto-encoder, respectively.
The orthogonal common space enables the classes to be more discriminative, thus the model can achieve better performances.
In addition,  PQZSL compresses the visual features into compact codes using quantizers to approximate the nearest neighbors.
In adition, study~\cite{shao2020generalized} adopts two variational auto-encoder (VAE) models to learn the cross-modal latent features from the visual and semantic spaces, respectively.
Cross alignment and distribution alignment strategies are devised to match the features from different spaces.


\vspace{-0.2cm}
\subsubsection{Other Methods}
\label{Sec:sec:others}
Besides the aforementiond categories, several studies employ various strategies to tackle GZSL problems.
In~\cite{das2019zero,Cacheux2019modeling,jiang2019transferable,zhu2019semantic}, the devised methods take into account the inter-class and intra-class relations among different classes.
Das and Lee~\cite{das2019zero} minimized the discrepancy between the semantic representations and visual features using the least square loss method. 
Rational matrices are constructed for each space, in order to minimize the inter-class pairwise relations between two spaces.
To avoid the projection domain shift problem, a point-to-point correspondence between the semantic representations and test samples is found.
The \textit{transferable constructive network} (TCN)~\cite{jiang2019transferable} consists of two parts: information fusion and constructive learning.
For information fusion, both visual features and semantic representations are encoded into a latent space.
Constructive learning checks how well an image is consistent with respect to a class by considering two aspects.
The first is whether learning is discriminative enough to recognize different classes, which uses semantic representations of the seen classes for supervision.
The second is whether learning is transferable to the unseen classes based on the class similarity score between visual features of the seen and unseen classes.
In~\cite{jin2019discriminant}, the joint optimization of center loss and softmax loss functions are adopted to learn more discriminative visual features for different classes while minimizing the intra-class variations.
Besides that, the performance of GZSL can be improved by rectifying the model output.\par


The studies in~\cite{jiang2018learning,wang2020learning} devise a dictionary-based framework for GZSL.
The model proposed in~\cite{jiang2018learning} jointly aligns the visual-semantic structure to constructs a class structure between the visual and semantic spaces by obtaining the class prototypes in both spaces. 
The aim is to explore some bases in each space to represent each class.
A domain adaptation is proposed to learn the prototypes from both seen and unseen classes in the visual space, in order to alleviate the problem of the projection domain shift.
On the other hand, study~\cite{wang2020learning} incorporates a sparse coding technique to construct a dictionary for each projection, i.e., visual-latent and semantic-latent, and applies an orthogonal projection to make the model discriminative.\par

Since the information obtained from the semantic representations, in particular human-defined attributes, is limited and less discriminative, recognizing data samples from different classes in a specific domain is difficult.
The studies in~\cite{zhang2019co,jin2019beyond,jiang2018learning,li2020ajoint} aim to address this issue.
The study~\cite{zhang2019co} decomposes the semantic vectors of the training set into $K$ subsets using the $k$-means clustering algorithm, and projects them to the visual space.
This decomposition allows the model to construct a uniform embedding space with a large local relative distance.
Similar to~\cite{zhang2019co}, Li et al.~\cite{li2020transferrable} used super-classes, which are generated by data-driven clustering, across the seen and unseen class domains to tune two domains.
Jin et al.~\cite{jin2019beyond} projected the semantic vector of each class to a high-order attribute space by applying the Gaussian random projection.
The model proposed in~\cite{li2020ajoint} defines a discriminative label space and projects the visual features via a linear projection matrix to that space.
Specifically, it fixes the labels of the seen classes and uses the attributes to map the label of the unseen classes into the label space.\par

\vspace{-0.35cm}
\subsection{Generative-based Methods}
\label{Sec:sec:gen}
{\color{black}Generative-based methods are originally designed to generate examples from the existing ones to train DL models~\cite{denton2015deep} and compensate the imbalanced classification problems~\cite{Guo2004learning}.}
Recently, these methods have been adopted to generate samples (i.e., images or visual features) for unseen classes by leveraging their semantic representations.
The generated data samples are required to satisfy two conflicting conditions: \emph{(i)} semantically related to the real samples, and \emph{(ii)} discriminative so that the classification algorithm can classify the test samples easily.
To satisfy the former condition, some underlying parametric representations can be used, while the classification loss function can be used to satisfy the second condition~\cite{bucher2017generating}.
Generative adversarial networks (GANs)~\cite{goodfellow2014generative} and variational autoencoders (VAEs)~\cite{kingma2013auto} are two prominent members of generative models that have achieved good results in GZSL.
{\color{black} In the following sub-sections, a review on generative-based methods under the semantic transductive learning setting is presented.
A summary of this category is provided in Table~\ref{Table:gen}.\par

}

\subsubsection{Generative Adversarial Networks}
\label{Sec:sec:sec:gan}
GANs generate new data samples by computing the joint distribution $p(y,x)$ of samples utilizing the class conditional density $p(x|y)$ and class prior probability $p(y)$.
GANs consist of a generator $G_{SV}: \mathcal{Z} \times \mathcal{A}\to\mathcal{X}$ that uses semantic attributes $\mathcal{A}$ and a random uniform or Gaussian noise $z\in \mathcal{Z}$, to generate visual feature $\tilde{x}\in \mathcal{X}$, and a discriminator $D_V:\mathcal{X}\times \mathcal{A}\to [0,1]$ that distinguishes real visual features from the generated ones.
When a generator learns to synthesize data samples for the seen classes conditioned on their semantic representations $\mathcal{A}_s$, it can be used to generate data samples for the unseen classes through their semantic representations $\mathcal{A}_u$.
However, the original GAN models are difficult to train, and there is a lack of variety in the generated samples.
In addition, the mode collapse is a common issue in GANs, as there are no explicit constraints in the learning objective.
To overcome this issue and stabilize the training procedure, many GAN models with alternative objective functions have been developed.
Wasserstein GAN (WGAN)~\cite{arjovsky2017wasserstein} mitigates the mode collapse issue using the Wasserstein distance as the objective function.
It applies a weight clipping method on the discriminator to allow incorporation of the Lipschitz constraint.
The improved WGAM model~\cite{gulrajani2017improved} reduces the negative effects of WGAN by utilizing gradient penalty instead of weight clipping.\par

\begin{figure}
    \includegraphics[width=0.47\textwidth]{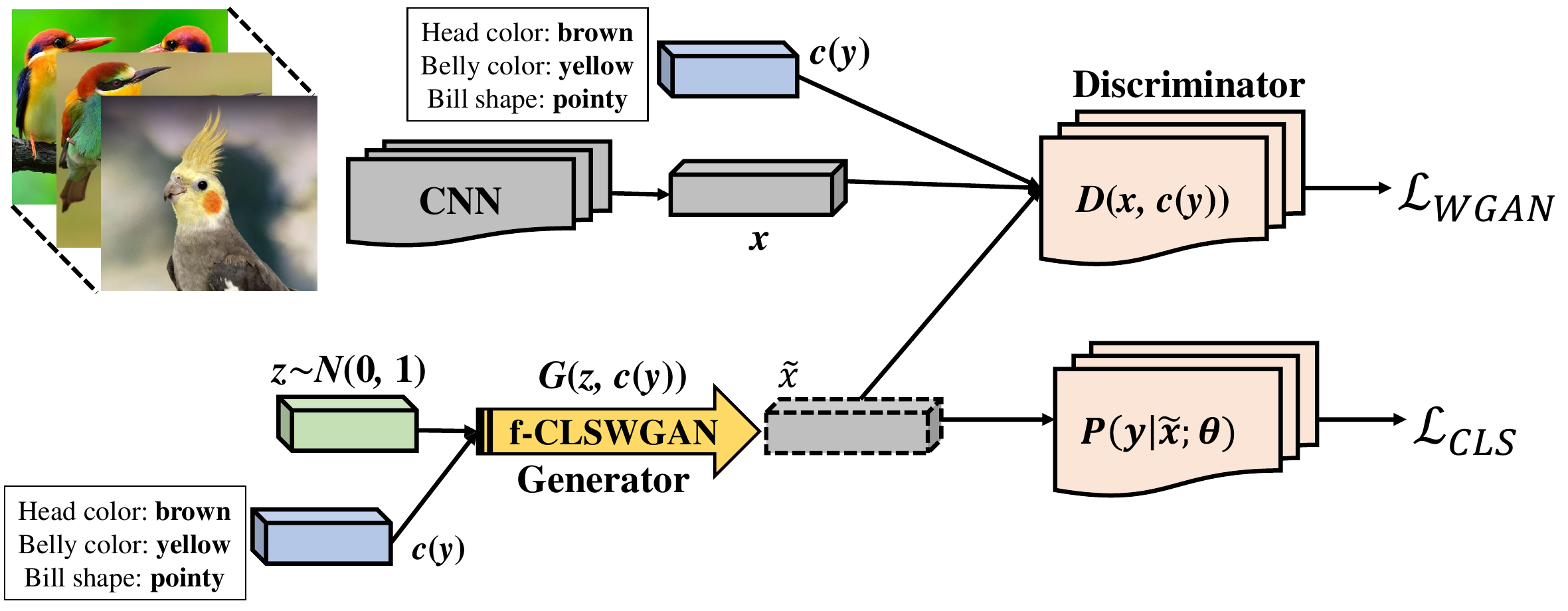}
    \vspace*{0.15cm}
    \caption{\color{black} A general view of the f-CLSWGAN model~\cite{xian2018feature}). It learns a generative model based on the visual features of seen classes conditioned on their corresponding semantic representations. It also uses classification loss to generate more discriminative visual features. }
    \label{f-CLSWGAN}
\end{figure}

Xian et al.~\cite{xian2018feature} devised a conditional WGAN model with a classification loss to synthesize visual features for the unseen classes, which is known as \emph{f-CLSWGAN} (see Fig.~\ref{f-CLSWGAN}).
To synthesize the related features, the semantic feature is integrated into both generator and discriminator by minimizing the following loss function:
\begin{multline} \label{eq:losswgan}
\mathcal{L}_{WGAN}=E[D(x^s,a^s)]-E[D(\tilde{x}^s,a^s)]\\-
\lambda E[(\|\bigtriangledown_{\hat{x}}D(\hat{x},a^s)\|_2-1)^2],
\end{multline}
where $\tilde{x}^s=G(z,a^s)$, $\hat{x}=\alpha x^s+(1-\alpha)\tilde{x}^s$ with $\alpha\sim U(0,1)$, $\lambda$ is penalty factor and discriminator $D:\mathcal{X}\times \mathcal{C} \to \mathbb{R}$ is a multi-layer perceptron.
In~(\ref{eq:losswgan}), the first and second terms approximate the Wasserstein distance, while the last term is the gradient penalty.
In addition, to generate discriminative features, the negative log-likelihood is used to minimize the classification loss, i.e.,:
\begin{align} \label{eq:loglikelihood}
\mathcal{L}_{CLS}=-E_{\tilde{x}^s\sim p_{\tilde{x}^s}}[\log P(y^s|\tilde{x}^s;\theta)],
\end{align}
where $P(y^s|\tilde{x}^s;\theta)$ is the probability that $\tilde{x}^s$ is predicted as class $y^s$, which is estimated by a softmax classifier by $\theta$ and optimized by visual features of the seen classes.
The final objective function can be written as:
\begin{align} \label{eq:obj}
\min_G \max_D \mathcal{L}_{WGAN}+\beta \mathcal{L}_{CLS},
\end{align}
where $\beta$ is the weighting factor.\par

{\color{black}

Since then, the conditional GAN (CGAN) has been combined with different strategies to generative discriminative visual features for the unseen classes, including \textit{optimal transport}-based approach~\cite{wang2021zero}, TFGNSCS~\cite{lin2019transfer} which is an extended version of f-CLSWGAN that considers transfer information, meta-learning approach~\cite{verma2020meta} which is based on \textit{model-agnostic meta-learning}~\cite{finn2017model}, LisGAN~\cite{li2019leveraging} which focuses on extracting information from soul samples, SPGAN~\cite{ma2020simila} which devises a similarity preserving loss with classification loss, \textit{Semantic rectifying GAN} (SR-GAN)~\cite{ye2019sr} which employs the semantic rectifying network (SRN)~\cite{maas2013rectifier} to rectify features, CIZSL~\cite{elhoseiny2019creativity,elhoseiny2021cizsl} which is inspired by the human creativity process~\cite{martindale1990clockwork}, \textit{MGA-GAN}~\cite{xie2021generalized} which uses multi-graph similarity, and MKFNet-NFG~\cite{xiang2021multi} which is adaptive fusion module based on the attention
mechanism.
Xie et al.\cite{xie2021cross} proposed cross knowledge learning and taxonomy regularization to train more relevant semantic features and generalized visual features, respectively.
In addition, Bucher et al.~\cite{bucher2017generating} developed four different conditional GANs, i.e., generative moment matching network (GMMN)~\cite{li2015generative}, AC-GAN~\cite{odena2017conditional}, denoising auto-encoder~\cite{bengio2013generalized}, and adversarial auto-encoder~\cite{makhzani2015adversarial}, to generate data samples for the unseen classes.
The empirical results indicate that GMMN outperforms other methods.\par

However, the generative-based models synthesize highly the unconstrained visual features for the unseen classes that can produce synthetic samples far from the actual distribution of real visual features.
To alleviate this issue and address the unpaired training issue during visual feature generation, Verma et al.~\cite{kumar2018generalized} proposed a cycle consistency loss (which is discussed further in the next sub-section), to ensure the generated visual features map back into their respective semantic space. Using this feedback mechanism allows the model to generate more discriminative visual features. In addition, the reconstructing generated features, which are unlabeled, are useful for the model to operate as a semi-supervised setting.\par

Later, the studies in~\cite{jian2019dual,zhu2017unpaired,chen2020canzsl,felix2018multi,li2019alleviating,li2021investigating} devise a cycle consistency loss term in their objective functions.
Specifically, DASCN~\cite{jian2019dual}, which is a dual learning model, combines a classification loss function with a semantics-consistency adversarial loss function. In the \textit{cycle-consistent adversarial network} for ZSL (CANZSL)~\cite{chen2020canzsl}, visual features are first synthesized from noisy text.
Then an inverse adversarial network is adopted to convert the generated features into text, in order to ensure that the synthesized visual features accurately reflect the semantic representations.
In addition, in studies~\cite{felix2018multi,li2019alleviating,li2021investigating}, a multi-modal cycle consistency loss function is developed to preserve the semantic consistency of the generated visual features.
AFC-GAN~\cite{li2019alleviating} introduces a boundary loss function to maximize the decision boundary between the seen and unseen features.
Boomerang-GAN~\cite{li2021investigating} uses a bidirectional auto-encoder to judge the reconstruction loss of the generated features and semantic embedding.
Moreover, \textit{two-level adversarial visual-semantic coupling} (TACO)~\cite{chandhok2021two} maximizes the joint likelihood of visual and semantic features by augmenting a generative network with inference.
This joint learning allows the model to better capture the underlying modes of the data distribution.\par

}

\vspace{-0.35cm}
\subsubsection{Variational Autoencoders (VAEs)}
\label{Sec:sec:sec:var}
VAE models identify the relationship between data sample $x$ and the distribution of the latent representation $z$.
A parameterized distribution $q_{\Phi}(z|x)$ is derived to approximate the posterior probability $p(z|x)$.
Similar to GAN, VAE models consist of two components: \emph{(i)} an encoder $q_{\Phi}(z|x)$ with parameter $\Phi$, and \textit{(ii)} a decoder $p_{\theta}(x|z)$ with parameter $\theta$.
The encoder maps sample space $x$ to latent space $z$ with respect to its class $c$, while a decoder maps the latent space back to the sample.
Conditional VAE (CVAE)~\cite{sohn2015learning} synthesizes sample $\hat{x}$ with certain properties by maximizing the lower bound of conditional likelihood $p(x|a)$, i.e.,
\begin{multline} \label{eq:cva}
\mathcal{L}(\Phi,\theta;x,a)=-KL(q_{\Phi}(z|x,a)\|p_{\theta}(z|a))\\
+E_{q_{\Phi}(z|a)}[\log p_{\theta}(x|z,a)].
\end{multline}

\begin{figure}
    \includegraphics[width=0.47\textwidth]{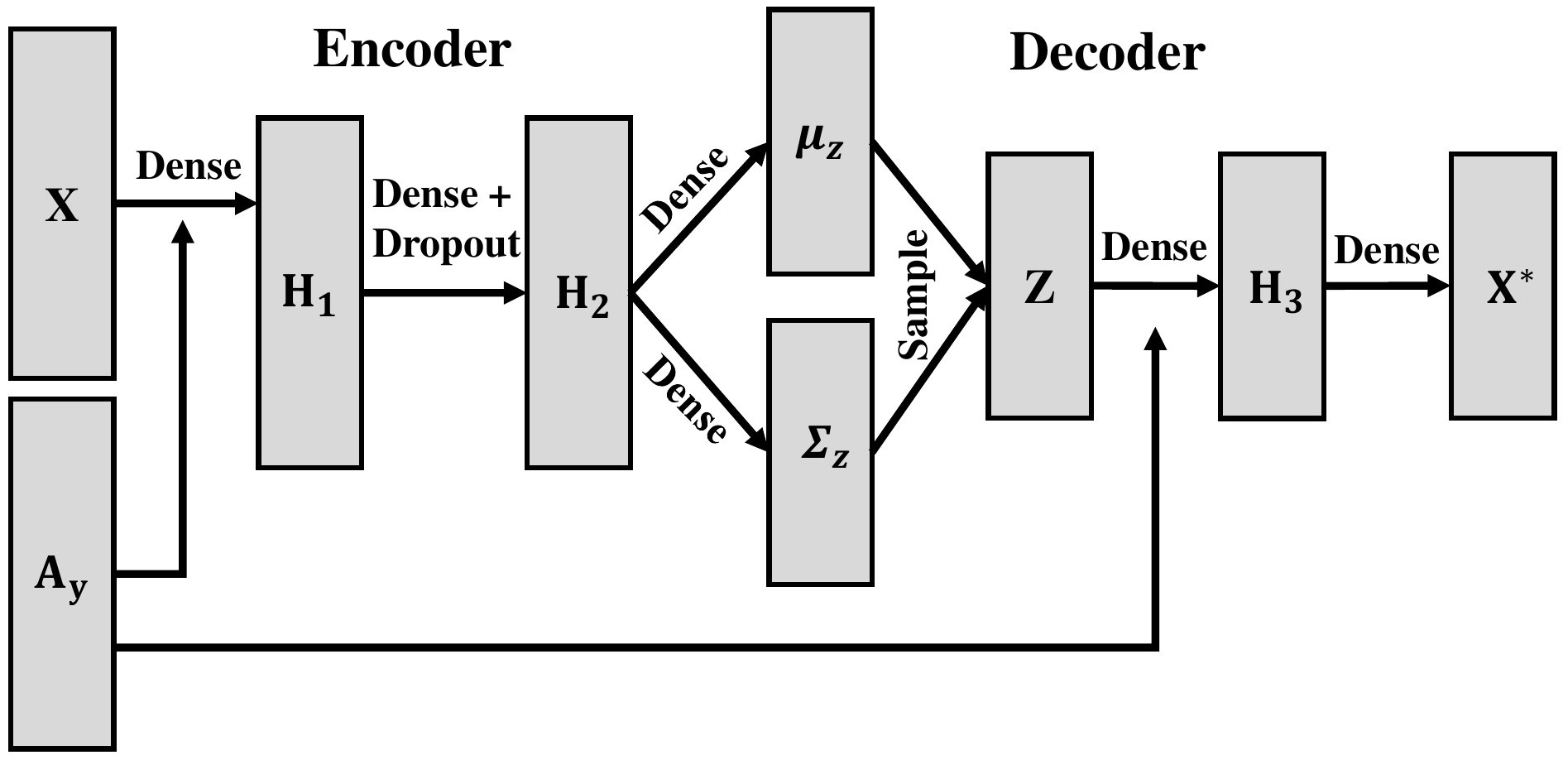}
    \vspace*{0.15cm}
    \caption{\color{black}A schematic view of CVAE-ZSL~\cite{mishra2018generative}. After concatenating the visual features and the semantic representations, it is passed through dense, dropout and dense layers. Then, another dense layer is used to output $\mu_z$ and $\sum_z$. Next, a $z$ is sampled from $\mathcal{N}(\mu_{x_i},\Sigma_{x_i})$ and projected to the image space for reconstruction.}
     \label{fig:CAVAE-ZSL}
\end{figure}

{\color{black} Studies~\cite{mishra2018generative,zhu2020don} use CVAE-based frameworks to generate visual features for GZSL.
CVAE-SZL~\cite{mishra2018generative} (Fig.~\ref{fig:CAVAE-ZSL}) adopts a neural network to model both encoder and decoder.
During training, encoder computes $q(z|x_i,A_{y_i})=\mathcal{N}(\mu_{x_i},\Sigma_{x_i})$ for each training sample $x_i$.
Then, $\mathcal{N}(\mu_{x_i},\Sigma_{x_i})$ is applied to sample $\widetilde{z}$.
Next, decoder is employed to reconstruct $x$ using $\widetilde{z}$ and $a_y$ with
the following loss function:
\begin{multline} \label{eq:cvaee}
\mathcal{L}(\Phi,\theta;x,A_y)=\mathcal{L}_{reconstr}(x,\hat{x})\\+KL(\mathcal{N}(\mu_x,\Sigma_x),\mathcal{N}(0,I)),
\end{multline}
where $\hat{x}$ represents the reconstructed sample and $L_2$ norm indicates the reconstruction loss.\par


In addition, several bi-directional CVAE methods map the generated visual features back to their semantic features to preserve the consistency of both features and produce high-quality examples for unseen classes.
In this regard, Verma et al.~\cite{kumar2018generalized} devised a cycle consistency loss function.
Their method, i.e., SE-GZSL, is equipped with a discriminator-driven feedback mechanism that maps a real sample $x$ or generated sample $\hat{x}$ back to the corresponding semantic representation.
The overall loss function of this discriminator is as follows:
\begin{align}
    \min_{\theta_R}\mathcal{L}_R=\mathcal{L}_{Sup}+\lambda_R\cdot\mathcal{L}_{Unsup}
\end{align}
where $\mathcal{L}_{Sup}$ is a supervised loss that learns from labeled examples, i.e.,
\begin{align}
    \mathcal{L}_{Sup}(\theta_R)=-E_{x_s,a_s}[p_{\theta_R}(a_s|x_s)],
\end{align}
while $\mathcal{L}_{Unsup}$ is an unsupervised loss function that learns from the unlabeled generated features $\hat{x}$, i.e.,
\begin{align}
    \mathcal{L}_{Unsup}(\theta_R)=E_{p_{\theta_G}(\hat{x}|z,a)p(z)p(a)}[p_{\theta_R}(a|\hat{x})].
\end{align}

Another CVAE-based bidirectional learning model is GDAN~\cite{huang2019generative}.
It contains a discriminator to estimate the similarity level with respect to each visual-textual feature pair.
The discriminator communicates with two other networks using a dual adversarial loss function.
CADA-VAE~\cite{schonfeld2019generalized} learns the shared cross-modal latent features of image and class attributes through a set of distribution alignment (DA) and cross alignment (CA) loss functions.
A VAE model~\cite{kingma2013auto} is exploited to learn the latent features of each data modality, i.e., image and semantic representations.
Then, the CA loss function, which can be estimated through decoding the latent feature of a data sample from other modality of the same class, is used for reconstruction.
In addition, the DA loss function is used to match generated image and class representations by minimizing their distance.
Chen et al.~\cite{chen2021entropy} attempted to minimize the uncertainty in the overlapped areas of both seen and unseen classes using an entropy-based calibration. Li et al.~\cite{li2019generalized} employed a multivariate regressor to map back the output the VAE decoder to the class attributes.
}

\subsubsection{Combined GANs and VAEs}
\label{Sec:sec:sec:genvar}
On one hand, VAEs generate blurry images due to the use of the element-wise distance in their structures~\cite{larsen2016autoencoding}.
On the other hand, the training process of GANs is not stable~\cite{salimans2016improved}.
{\color{black}To address these limitations, Larsen et al.~\cite{larsen2016autoencoding} proposed VAEGAN, in which the discriminator in GAN is used to measure the similarity metric.
VAEGAN with similarity measure is able to generate better samples than models with element-wise metrics.
Later, Gao et al.~\cite{gao2020zerovaegan,gao2018joint} proposed a joint generative model (known as Zero-VAE-GAN) to generate high-quality visual features for the unseen classes. 
More specifically, CVAE conditioned on semantic attributes is combined with CGAN conditioned on both categories and attributes.
Besides that, a categorization network and a perceptual reconstruction loss~\cite{johnson2016perceptual} are incorporated to generate high-quality features.}
Verma et al.~\cite{verma2021towards} leveraged the meta-learning strategy to train a generative model that integrates CVAE and CGAN to generate visual features from data sets that contain few samples for each class.
Moreover, Xu et al.~\cite{xu2021dual} proposed a dual learning framework based on CVAE and CGAN with an additional classifier to generate more discriminative visual features for unseen classes.\par

\subsubsection{Other methods}
\label{Sec:sec:sec:other}
Apart from GANs and VAEs, several studies have attempted to generate visual features for the unseen classes using other approaches.
{\color{black}As an example, the \textit{unseen visual data synthesis} (UVDS)~\cite{long2018zero} method exploits a diffusion regularization (DR) to synthesize visual features for the unseen classes using embedding matrices.
The \textit{class-specific synthesized dictionary} (CSSD)~\cite{Ji2019class} learns a class-specific encoding matrix in a latent space for each class and consequently a dictionary matrix within a dictionary framework. Then, the encoding matrices with the affinity seen classes, i.e., seen classes similar to unseen ones, are used to generate visual features for the unseen classes.
Feng and Zhao~\cite{feng2020transfer} built a generative model by extracting two types of knowledge, i.e., local rational knowledge and global rational knowledge, from the visual and semantic representations, respectively.
Li et al.~\cite{li2020learning} leveraged the most similar seen classes to the unseen ones in the semantic space to generate visual features for the unseen classes.\par

In~\cite{shi2020discriminative,yu2018bi,ji2019multi,liu2021task,luo2020generative}, autoencoders are used to generate visual features for the unseen classes.
Shi and Wei~\cite{shi2020discriminative} developed an encoder to map the visual features to the semantic embedding space.
The regressor feedback is imposed into the decoder to reconstruct truthful visual features.
Then, the learned decoder generates visual features for the unseen classes to train a classifier.
{\color{black}\emph{Bi-adversarial auto-encoder} (BAAE)~\cite{yu2018bi} pairs an autoencoder with two adversarial networks.
On one hand, the encoder, which operates as a generator, integrates the visual and synthesized features into an adversarial network to capture the real distribution.
On the other hand, the decoder, which acts as semantic inference, formulates real class semantics for inference toward another adversarial network to enforce both real and synthesized visual features to be related to the semantic representation.
Similar to BAAE, \textit{the multi-modality adversarial auto-encoder} (MAAE)~\cite{ji2019multi} pairs an auto-encoder with multi-modality adversarial networks. The encoder generates visual features while the decoder aims to relate both generated and real features to the class semantics.
Both BAAE and MAAE integrate classification networks to ensure that the generated and real features are discriminative.}
Moreover, Liu et al.~\cite{liu2021task} tackled the limitations caused by diverse data distribution in GZSL by proposing a meta-learning framework based on autoencoders.\par

}


In contrast, the studies in~\cite{changpinyo2020classifier,changpinyo2017predicting,yu2020episode} attempt to synthesize classifiers instead of generating visual features for the unseen classes.
EXEM~\cite{changpinyo2020classifier,changpinyo2017predicting} learns a function to predict the locations of visual features with respect to the unseen classes. 
The visual exemplar of each class is created via averaging the Principal Component Analysis (PCA) projection of the data samples belonging to a particular class.
Then, $D$ regressors are learned with respect to the visual exemplar semantic representations.
Finally, the similarity level between the test samples and bases is computed with regard to the predicted exemplars to produce a prediction.
E-PGN~\cite{yu2020episode}, which is an episode-based framework, generates class-level visual samples conditioned on semantic representations.

{\color{black}
\vspace{-0.3cm}
\section{Transductive GZSL Methods }
\label{Sec:sec:trans}
As explained earlier, GZSL methods under the transductive learning setting can alleviate the projection domain shift problem by taking unlabeled data samples from the unseen classes into consideration.
Accessing the unlabeled data allows the model to know the distribution of the unseen classes and consequently learn a discriminative projection function.
It also permits synthesis of the related visual features for the unseen classes by using generative based methods.
Since the number of publications is limited, we categorize transductive-based GZSL methods into two: embedding-based and generative-based methods.
Table~\ref{Table:trans} summarizes the transductive-based GZSL methods.\par

\vspace{-0.2cm}
\subsection{Embedding-based methods}
\label{Sec:sec:trans-embd}
This category of transductive-based GZSL methods leverages the unlabeled data samples from the unseen classes to learn a projection function between the visual and semantic spaces.
Using these unlabeled data samples, the geometric structure of the unseen classes can be estimated.
Then, a discriminative projection function in the common space, i.e., semantic, visual, or latent, can be formulated, in order to alleviate the projection domain shift problem.
To map the visual features of the seen classes to their corresponding semantic representations, a classification loss function is derived, and an additional constraint is required to extract useful information from the data samples from the unseen classes.

\begin{figure}[tb]
    \centering
    \includegraphics[width=0.47\textwidth]{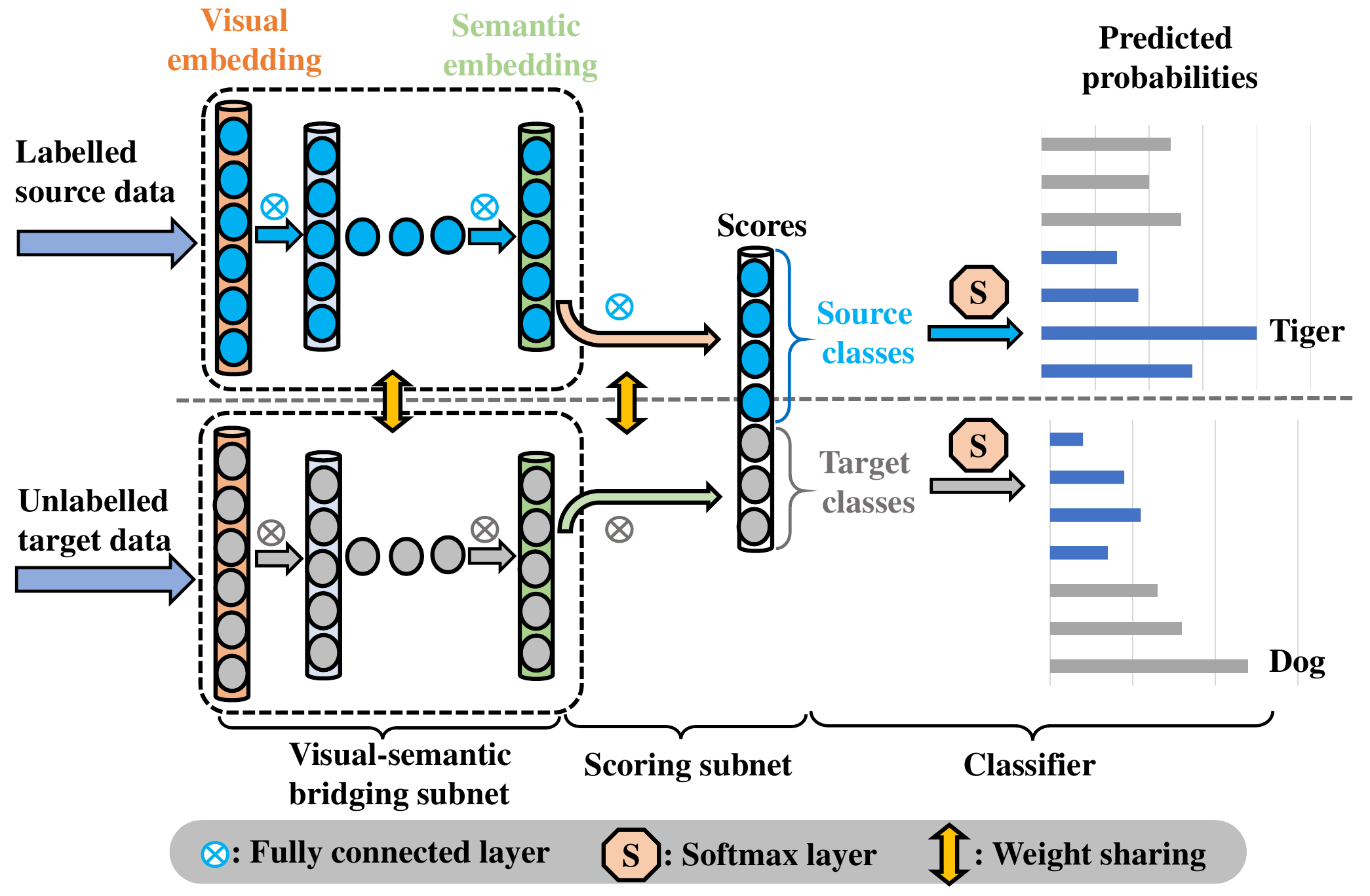}
    \vspace*{0.15cm}
    \caption{\color{black}A general view of QFSL~\cite{song2018transductive}. It first projects the visual features of the seen classes into a number of fixed points in the semantic space. Then, a bias loss function~(\ref{eq:qfsl1}) is used to map the visual features of the unseen classes into other points. }
    \label{QFSL}
\end{figure}

An example is \textit{Quasi-fully supervised learning} (QFSL)~\cite{song2018transductive} (Fig.~\ref{QFSL}), which projects the visual features of the seen classes into a number of fixed points in a semantic space using a fully connected network with the ReLU activation function.
The unlabeled visual features from the unseen classes are mapped onto other points with the following loss function.
\begin{align} \label{eq:qfsl}
\mathcal{L}=\frac{1}{N_s}\sum_{s=1}^{N_s}\mathcal{L}_{p}(x_s)+\frac{1}{N_u}\sum_{u=1}^{N_u}\gamma \mathcal{L}_{b}(x_u)+\lambda \Omega (W),
\end{align}
where $\mathcal{L}_{p}$ and $\Omega$ are the classification loss function and regularization factor, $\mathcal{L}_{b}$ is the bias loss, i.e.,
\begin{align} \label{eq:qfsl1}
\mathcal{L}_{b}=-\ln{\sum _{u\in Y^u}p_u},
\end{align}
\vspace*{-0.035cm}
where $p_u$ indicates the predicted probability of the $u$-th class.\par

\begin{table*}[th!]\color{black}
\centering
\caption{\label{Table:data} The details of data sets.}
    \begin{adjustbox}{width=0.99\textwidth}
    \begin{tabular}{l c c c c c c c c c c c}
    \toprule
   Data set & \# of seen/unseen classes& Semantic vectors/dimension & \# of training/ testing samples  \\
      \midrule
      CUB & 150/50 & attributes/312 & 7507/1764+2967\\
      AWA1 & 40/10 & attributes/85 & 19832/4958 + 5685\\
      AWA2 & 40/10 & attributes/85 & 23527/5882 + 7913\\
      SUN & 645/72 & attributes/102 & 10320/2580 + 1440\\
      aPY & 20/12 & attributes/64 & 5932/1483 + 7924\\
      NAB & 323/81 & descriptions/13217 & 48562/NA \\
      DeepFashion & 36/10 & attributes/1000 & 204885/84337 \\
      LADdataset & 184/46 & attribute/359 & 54610/23407\\
      Dogs & 85/28 & descriptions/NA& 19501/NA\\
      FLO & 82/20 & sentences/1024 & 5631/1403+1155\\
    \bottomrule
     \end{tabular}
    \end{adjustbox}
\end{table*}

Other semantic embedding models have been proposed, e.g.~\cite{fu2018zero,niu2018zero} and~\cite{li2017zero}.
Fu et al.~\cite{fu2018zero} developed a semantic manifold structure of the class prototypes distributed in the embedding space.
A \textit{dual visual semantic mapping paths} (DMaP)~\cite{li2017zero} is formulated to learn the connection between the semantic manifold structure and visual-semantic mapping of the seen classes.
It extracts the inter-class relation between the seen and unseen classes in both spaces.
For a given test sample, if the inter-class relationship consistency is satisfied, a prediction is produced.
On the other hand, \textit{adaptive embedding ZSL} (AEZSL)~\cite{niu2018zero} learns a visual-semantic mapping for each unseen class by assigning higher weights to the recognition tasks pertaining to more relevant seen classes.\par

Several studies~\cite{guan2018extreme,huo2018zero,zhang2020deep} focus on the development of visual embedding-based models under the transductive setting.
Hu et al.~\cite{huo2018zero} leveraged super-class prototypes, instead of the seen/unseen class prototypes, to align with the seen and unseen class domains.
The $K$-means clustering algorithm is used to group the semantic representations of all seen and unseen classes into $r$ superclass.
The DTN~\cite{zhang2020deep}, which is a probabilistic-based method, decomposes the training phase into two independent parts.
One phase adopts the cross-entropy loss function to learn the labeled seen classes, while the other part applies a combination of cross-entropy loss function and Kullback–Leibler divergence to learn from the unlabeled samples in the unseen classes.
In the second phase, cross-entropy is adopted to avoid the transferring of samples from the unseen classes into seen classes.\par

MFMR~\cite{xu2017matrix} employs a matrix tri-factorization~\cite{ding2006orthogonal} framework to construct a projection function in the latent space.
Two manifold regularization terms are formulated to preserve the geometric structure in both spaces, while a test time manifold structure is developed to alleviate the shift problem.
In~\cite{long2018pseudo,zhang2020towards}, the pseudo labeling technique is used to solve the bias problem by incorporating the testing samples into the training process.\par

\vspace{-0.2cm}
\subsection{Generative-based methods}
\label{Sec:sec:trans-gen}
This category of GZSL methods uses the unlabeled unseen data samples to generate the related visual features for the unseen classes.
A generative model is usually formulated using the data samples of the seen classes, while the unlabeled samples from the unseen classes are used to fine-tune the model based on unsupervised strategies~\cite{gao2020zerovaegan,wu2020self}.
As an example, Gao et al.~\cite{gao2020zerovaegan} developed two self-training strategies based on pseudo-labeling procedure.
The first strategy uses $K$-nearest neighbour (K-NN) algorithm to provide pseudo-labels for the unseen visual features.
As such, the semantic representations of the unseen classes are used to generate $N$ fake unseen visual features from the Gaussian distribution.
Then, for each unseen class, the average of $N$ fake features is used as an anchor.
At the same time, $K$-NN is used to update the anchor.
Finally, the top $M$ percent of unseen features are selected to fine-tune the generative model.
The second strategy obtains the pseudo-labels directly through the classification probability.
SABR-T~\cite{paul2019semantically} uses WGAN to generate the latent space for the unseen classes via minimizing the marginal difference between the true latent space representation of the unlabeled samples of unseen classes and the generated space.
On the other hand, VAEGAN-D2~\cite{xian2019f} learns the marginal feature distribution of the unlabeled samples using an additional unconditional discriminator.\par

\vspace{-0.2cm}
\section{Applications}
\label{Sec:app}
Due to the ubiquitous demand of machine learning techniques for large scale data sets and the rapid digital advances in recent years, the GZSL methods are now being applied to a variety of applications, particularly in computer vision and natural language processing (NLP).
We discuss these applications in the following subsections, which are divided into image classification (the most popular application of GZSL), object detection, video processing, and NLP.\par
\vspace{-0.2cm}
\subsection{Computer vision}
\label{Sec:sec:image}
In computer vision, GZSL is applied to solve problems related to both images and videos.

\begin{table*}[tb] \color{black}
\centering
\caption{\label{Table:com}\color{black} Highlights of GZSL methods. }
   \begin{adjustbox}{width=0.99\textwidth}

    \begin{tabular}{l c p{0.5\textwidth}}
    \toprule
     {Methods} & {Description} \\

    \midrule
       {Embedding}&  {\begin{minipage}{3.7in}Less complex structure and easy to implement; various projection functions such as linear and non-linear according to the properties of the data set can be selected. However, these methods suffer from hubness, shift and bias problems when a semantic embedding space is learned, as well  as shift and bias problems when a visual or latent space is learned. \end{minipage} }      \\
    \cmidrule{1-2}
     {Generative} & {\begin{minipage}{3.7in}Large number of samples can be generated for the unseen classes; a variety of supervised learning models can be used to perform recognition. However, they are complex in structure, difficult to train, unstable in training, and susceptible to the mode collapse issue.\end{minipage} } \\
     \bottomrule
     \end{tabular}
   \end{adjustbox}
\end{table*}

\textbf{Image processing:}
\textit{Image classification} is among the most popular applications of GZSL methods. The aim is to classify images from both seen and unseen classes.
In this regard, ImageNet~\cite{imahenet2009deng}, which is a WordNet hierarchy~\cite{miller1995wordnet} data set, is one of the well-known image data sets that is widely used in various studies with different settings in the computer vision and image processing domain. The original ImageNet includes over 15 million labeled images (high-resolution) related to approximately 22,000 classes (categories). Both fine-grained (sub-ordinate classes, e.g., various vehicle types or birds) and coarse-grained (without sub-ordinate classes, e.g. table, animal and bus) data samples are available. Several researchers used the full ImageNet data set in their studies, e.g.~\cite{norouzi2013zero,changpinyo2016synthesized}. On the other hand, Rohrbach et al.~\cite{rohrbach2011evaluating} used a balanced subset of ImageNet with 1000 classes to evaluate their proposed GZSL model, where 800 classes were considered as seen classes while the remaining 200 were unseen classes. Xian et al.~\cite{xian2017zero} used 1000 classes for training and the remaining 21,000 (and subsets of them) as the unseen classes. Word2Vec~\cite{mikolov2013distributed} trained on Wikipedia provided in~\cite{elhoseiny2017link} was utilised to obtain semantic information with respect to the ImageNet data set.\\

On the other hand, there exist several attribute-based data sets. The aPascal-aYahoo (aPY)~\cite{farhadi2009describing} and Animal with Attribute 1 (AWA1)~\cite{lampert2013attribute} are two coarse-grained data sets. The aPY~\cite{farhadi2009describing} is a small-scale data set containing 12,695 images pertaining to the original Pascal VOC 2008 data samples categorized into 20 different classes (aPascal). It also includes 2,644 images collected using the Yahoo image search engine (aYahoo) related to 12 classes. Then, these 20 Pascal classes and 12 Yahoo classes are used for training (seen classes) and test (unseen classes) purposes, respectively.  All classes are annotated by 64 binary attributes for specifying the visible objects. The Animal with Attribute (AWA)~\cite{lampert2009learning} contains over 30,000 animal images belonging to 50 classes. AWA1~\cite{lampert2013attribute} is a medium-scale coarse-grained version of AWA. It includes 30,475 images categorized into 40 classes for training and 10 additional classes for test, which covers 85 binary and continues attributes. AWA2~\cite{xian2017zero} is an extended version of AWA1. It has 37,322 images that do not overlap with those in the original AWA data set. It uses the same 50 classes of AWA1, with 85 attributes.  On average, each class has 746 images, in which ``mole" and ``horse" are the least and most populated classes with 100 and 1645 images, respectively.\par

Moreover, Scene UNderstanding (SUN)~\cite{xiao2010sun} and CUB-200-2011 (CUB)~\cite{wah2011the} data sets are two medium-scale and fine-grained data sets. SUN~\cite{xiao2010sun}, which is a well-known scenes related data set, contains 130,519 images with 899 classes. The SUN attribute data set~\cite{patterson2012sun, patterson2014sun} is a subset of the original SUN version for fine-grained scene classification. It has 14,340 images related to 717 classes with 102 attributes, where 645 classes including 12900 images (10,320 and 2580 images are, respectively, used as seen training and test samples) and the remaining 72 classes with 1440 images are used as unseen test samples. The attributes of SUN cover multiple wide areas such as ``trees” and ``manmade”, although limited images per class are available.
CUB~\cite{wah2011the} is an extended form of the original CUB-200 data set~\cite{welinder2010caltech}. It contains approximately two times more images in each class, as compared with those in the original CUB-200 version. All images in CUB-200-2011 are annotated with part locations, attribute labels, and bounding boxes. There are 11,788 images that belong to 200 different classes of birds, annotated with 312 attributes.  The attributes of CUB focus on local information, such as ``has wing pattern spotted” and ``has throat color orange”. Among them, 150 classes (containing 7057 images for training and 1764 images for test) are seen classes and the remaining 50 classes with 2967 images are unseen test samples.\par

North America Birds (NAB)~\cite{van2015building} is another fine-grained data that includes 48,562 images belonging to 1,011 species of commonly observed birds in North America. Elhoseiny et al.~\cite{elhoseiny2017link} extended the NAB data set with the corresponding unstructured text article for each class extracted from Wikipedia and re-organized it into 404 classes. A total of 323 classes are seen classes while the remaining 81 classes are unseen classes. Note that AWA2, SUN and CUB data sets have even distributions of samples per class.
In contrast, DeepFashion~\cite{liu2016deepfashion}, which is a large-scale data set of clothes with massive attributes of over 800,000 images, follows a more realistic long-tail distribution. It has 46 classes, i.e., 36 seen and 10 unseen classes, respectively.
Other available data sets pertaining to image classification tasks include Large-Scale Attribute Data set (LADdataset)~\cite{zhao2019large}, Dogs~\cite{khosla2011novel}, Oxford Flowers (FLO)~\cite{nilsback2008automated}.}
Table~\ref{Table:data} summarizes the details of the data sets.

\textit{Object detection} is another popular application in computer vision that has increasingly gained importance for large-scale tasks.
It aims to locate objects, in addition to their recognition.
Over the years, many CNN-based models have been developed to recognize the seen classes.
However, collecting sufficient annotated samples with ground-truth bounding-boxes is not scalable, and new methods have emerged.
Currently, advances in zero-shot detection allow the conventional object detection models to detect classes that do not match previously learned ones.
The reported methods in the literature include detecting general categories from single label~\cite{zhu2020dont,vyas2020zero,lee2018hierarchical,zablocki2019context,cheraghian2019mitigating,cacheux2021zero} to multi-label~\cite{lee2018multi,gupta2021generative}, multi-view (CT and x-ray )~\cite{Paul2021generalized} and tongue constitution recognition~\cite{wen2020grouping}.
In addition, GZSL has been applied to segment both seen and unseen categories~\cite{bucher2019zero,zhang2021zero}, retrieve images from large scale data sets~\cite{dutta2019generalized,zhu2020ocean} as well as perform image annotation~\cite{wang2019inductive}.

\textbf{Video processing: }
Recognizing human actions and gestures from videos is one of the most challenging tasks in computer vision, due to a variety of actions that are not available among the seen action categories.
In this regard, GZSL-based frameworks are employed to recognize single label~\cite{liu2019generalized,mishra2018generative,mandal2019out,gowda2021claster} and multi-label~\cite{wang2020multi} human actions.
As an example, {\color{black} CLASTER~\cite{gowda2021claster} is a clustering-based method using reinforcement learning.}
In~\cite{wang2020multi}, a multi-label ZSL (MZSL) framework using JLRE (\textit{Joint Latent Ranking Embedding}) is proposed.
The relation score of various action labels is measured for the test video clips in the semantic embedding and joint latent visual spaces.
In addition, a multi-modal framework using audio, video, and text was introduced in~\cite{parida2020coordinated,mazumder2021avgzslnet}.\par

\vspace{-0.32cm}
\subsection{Natural language processing}
\label{Sec:sec:NLP}
NLP~\cite{basiri2020novel,basiri2020abcdm} and text analysis~\cite{abdar2020energy}
are two important application areas of machine learning and deep learning methods.
In this regards, GZSL-based frameworks have been developed for different NLP applications such as text classification with single label ~\cite{wang2018zero,chen2020canzsl}, multi-label~\cite{song2020generalized, huynh2020ashared} as well as noisy text description~\cite{elhoseiny2017link,zhu2018agenerative}.
Wang et al.~\cite{wang2018zero} proposed a method based on the semantic embedding and categorical relationships, in which benefits of the knowledge graph is exploited to provide supervision in learning meaningful classifiers on top of the semantic embedding mechanism.\par

\begin{table*}[tb] \color{black}
\centering
\caption{\label{Table:comp}\color{black} Highlights of embedding and generative-based methods. }
   \begin{adjustbox}{width=0.99\textwidth}

    \begin{tabular}{l c c p{0.5\textwidth}}
    \toprule
     {Methods}& {Strategy} & {Description} \\

    \midrule
      \multirow{5}{*}{Embedding} &{Graph learning}&  {\begin{minipage}{3.6in}Able to preserves the geometric structure of features in the latent space, which leads to a compact representation; however, learning a classifier using graph information is challenging and they contain a complex structure. \end{minipage} }      \\
    \cmidrule{2-3}
     &{Meta learning} & {\begin{minipage}{3.6in}Able to learn discriminative features without labeled samples from novel classes even without fine-tuning.\end{minipage} } \\
      \cmidrule{2-3}
     &{Attention learning} & {\begin{minipage}{3.6in}Able to identify fine-grained classes and recognize important information pertinent to performing GZSL tasks; however, it increases computational load, affecting real-time implementation. \end{minipage} } \\
      \cmidrule{2-3}
     &{Bidirectional learning} & {\begin{minipage}{3.6in}Able to learn more generalizable projections. \end{minipage} } \\
       \cmidrule{2-3}
     &{Autoencoders} & {\begin{minipage}{3.6in}Able to learn in an unsupervised manner with a simple structure, which leads to more discriminative features and reduced structural complexity. \end{minipage} }  \\
     \midrule
     \multirow{5}{*}{Generative}&{GANs} & {\begin{minipage}{3.6in}Able to alleviate the bias and projection domain shift problems; however, GANs are difficult to train, and there is a lack of variety in the generated samples; in addition, they are susceptible to mode collapse and an unstable training process. \end{minipage} }  \\
     \cmidrule{2-3}
     & {VAEs} & {\begin{minipage}{3.6in} Able to alleviate the bias and projection domain shift problems; however,VAEs generate blurry images due to the use of the element-wise distance in their structures. \end{minipage} }  \\
     \bottomrule
     \end{tabular}
   \end{adjustbox}
\end{table*}

The application of GZSL on multi-label text classification plays a key role in generating latent features in text data. Song et al.~\cite{song2020generalized} proposed a new GZSL model in a study on International Classification of Diseases (ICD) to identify classification codes for diagnosis of various diseases. The model improves the prediction on the unseen ICD codes without compromising its performance on seen ICD codes.
On the other hand, for multi-label ZSL, Huynh and Elhamifar~\cite{huynh2020ashared} trained all applied models on the seen labels and then tested on the unseen labels, whereas for multi-label GZSL, they tested all models on the both seen and unseen labels.
In this regard, a new shared multi-attention mechanism with a novel loss function is devised to predict all labels of an image, which contain multiple unseen categories.\par

\vspace{-0.37cm}
\section{Discussions and Conclusions}
\label{Sec:dis}
In this section, we discuss the main findings of our review. Several research gaps that lead to future research directions are highlighted.

We categorize the GZSL methods into two groups: \textit(i) embedding-based, and \textit(ii) generative-based methods.
Embedding-based methods learn {\color{black}an embedding space}, whether visual-semantic, semantic-visual, common/latent space, or a combination of them (bidirectional), to link the visual space of the seen classes into their corresponding semantic space.
They use the learned {\color{black}embedding space} to recognize data samples from both seen and unseen classes.
{\color{black}Various strategies have been developed to learn the embedding space.
We categorize these strategies into graph-based, autoencoder-based, meta-learning-based, attention-based, compositional learning-based, bidirectional learning-based and other methods.
In contrast, the} generative-based methods convert GZSL into a conventional supervised learning problem by generating visual features for the unseen classes.
Since the visual features of the unseen classes are not available during training, learning a projection function or a generative model using data samples from the seen classes cannot guarantee generalization pertaining to the unseen classes.
Although the low-level attributes are useful for classification of the seen classes, it is challenging to extract useful information from structured data samples due to the difficulty in separating different classes.
Thus, the main challenge of both methods is the lack of unseen visual samples for training, leading to the bias and projection domain shift problems.
{\color{black}Table~\ref{Table:com} summarizes the main properties of the embedding- and generative-based methods. In addition, Table~\ref{Table:comp} describes different embedding- and generative-based methods. }\par

{\color{black}\textbf{\textit{Embedding-based vs. Generative-based methods:}}} Embedding-based models have been proposed to address the ZSL and GZSL problems.
These methods are less complex, and they are easy to implement.
However, their capabilities in transferring knowledge are restricted by semantic loss, while the lack of visual samples for the unseen classes causes the bias problem.
This results in poor performances of the models under the GZSL setting.
{\color{black}This can be seen in Table~\ref{Table:embeding}, where the accuracy of seen classes ($Acc_s$) are higher than accuracy of unseen classes ($Acc_u$).}
In semantic embedding models, the projection of visual features to a low dimensional semantic space shrinks their variance and restricts their discriminability~\cite{zhang2017learning}.
The compatibility scores are unbounded, and ranking may not be able to learn certain semantic structures because of the fixed margin. Moreover, they usually perform  search in the shared space, which causes the hubness problem~\cite{wu2019simple,liu2020label,chen2018zero}.

\begin{table*}[th!]   \color{black}
\centering
\caption{\label{Table:embeding} A summary of embedding-based methods (``S", ``V" and ``L" represent the semantic, visual and latent embedding space, respectively, and ``-" indicates the item is not reported by the corresponding study). }
    \begin{adjustbox}{width=0.99\textwidth}
    \begin{tabular}{l c c c c c c c c c c c}
    \toprule
   \multirow{2}{*} { Study} & Embedding & \multirow{2}{*} {Classifier} &\multicolumn{3}{c}{AWA2} & \multicolumn{3}{c}{CUB} & \multicolumn{3}{c}{SUN}  \\
      \cmidrule{4-12}
      & space&&$Acc_s$&$Acc_u$&$H$&$Acc_s$&$Acc_u$&$H$&$Acc_s$&$Acc_u$&$H$\\
      \midrule
      DCN~\cite{liu2018generalized}& L & Nearest neighbor &84.2 &25.5 &39.1 &60.7 &28.4 &38.7 &37.0 &25.5 &30.2 \\
      \cmidrule{2-12}
      DUET~\cite{jia2020deep} & S & Softmax &90.2  &48.2  &63.4  &80.1 &39.7 &53.1 &- &- &- \\
      \cmidrule{2-12}
      PSD~\cite{zhang2020pseudo}&V & Neural network &- &- &- &53.1 &38.6 &44.7 &39.7 &34.3 &36.8 \\
      \cmidrule{2-12}
      Zhu et al.~\cite{zhu2019generalized}& L & Neural network &88.7 &45.6 &60.2 &68.9 &39.5 &50.2 &- &- &- \\
      \cmidrule{2-12}
      Zhang et al.~\cite{zhang2018triple} & L & Least square or Sylvester equations &- &- &- &62.3 &26.5 &37.2 &38.3 &22.2 &28.1 \\
      \cmidrule{2-12}
      Self-focus~\cite{yang2019self}&V & -  &68.1 &39.0 &53.4 &47.5  &21.9 &30.0 &- &- &- \\
      \cmidrule{2-12}
      SABR-I~\cite{paul2019semantically}& L & Softmax &93.9 &30.3 &46.9 &58.7 &55.03 &56.8 &35.1 &50.7 &41.5 \\
      \cmidrule{2-12}
      Zhang et al.~\cite{zhang2020towards}&L & Nearest neighbor &93.2 &68.4 &78.9 &62.9 &54.0 &58.1 &38.5 &47.2 &42.4 \\
      \cmidrule{2-12}
      LAF~\cite{liu2020label} & S \& V & - &58.5 &50.4 &54.2 &52.0 &43.7 &47.5 &36.6 &36.0 &36.3 \\
      \cmidrule{2-12}
      SDM-Net~\cite{daghaghi2019semantic}  & S & Softmax &  78.6&  55.1& 64.7 &52.5 &47.1 &49.6 &32.6 &47.2 &38.6 \\
      \cmidrule{2-12}
      Rahman et al.~\cite{rahman2018aunified} & S & - &- &-  &-  &41.7 &44.9 &43.3 &27.8 &35.8 &31.3 \\
      \cmidrule{2-12}
      SP-AEN~\cite{chen2018zero}& S \& V & Neural network &90.9 &23.3 &37.1 &70.6 &34.7 &46.6 &38.6 &24.9 &30.3  \\
      \cmidrule{2-12}
      DSS~\cite{Yang2018dissimilarity} & L & Neural network &97.5 &15.7 &27.0 &93.2 &25.0 &39.4 &81.4 &18.2 &29.7\\
      \cmidrule{2-12}
      CRnet~\cite{zhang2019co}& V & Neural network&78.8  &52.6  &63.1  &56.8 &45.5 &50.5 &36.5 &34.1 &35.3 \\
      \cmidrule{2-12}
      APN~\cite{xu2020attribute} & S & - &69.5 &62.2 &65.6 &74.9 &65.7 &70.0 &39.2 &49.4 &43.7 \\
      \cmidrule{2-12}
      DARK~\cite{guo2019dual} & S \& V & similarity &- &- &38.3 &- &- &41.6 &- &- &32.8 \\
      \cmidrule{2-12}
      Das and Lee~\cite{das2019zero} & V& Nearest neighbor &72.3  &60.6  &65.9 &45.1 &44.0 &44.6 &36.6 &54.1 &43.7 \\
      \cmidrule{2-12}
      DAZLE~\cite{huynh2020fine}& S & Softmax &75.7  &60.3 &67.1  &59.6 &56.7 &58.1 &24.3 &52.3 &33.2 \\
      \cmidrule{2-12}
      AGZSL~\cite{wang2019asymmetric} & L & Nearest neighbor &68.5  &34.8  &46.1  &56.5 &19.4 &28.9 &28.6 &18.5 &22.5 \\
      \cmidrule{2-12}
      LSG~\cite{xu2021semi}& S & - &84.9 &60.4 &70.6 &50.4 &49.6 &50.0 &23.1 &52.8 &32.2  \\
      \cmidrule{2-12}
      Li et al.~\cite{li2020transferrable}& S \& V & Neural network &68.2 &44.5 &53.9 &46.9 &51.1 &48.9 &30.2 &31.4 &30.7 \\
      \cmidrule{2-12}
      DAGDA~\cite{li2020anchor} & L & - &91.5 &16.5 &28.0 &70.0 &23.5 &35.2 &31.0 &14.9 &20.1 \\
      \cmidrule{2-12}
      RGEN~\cite{xie2020region}& S & - &76.5 &67.1 &71.5 &68.5 &61.4 &64.7 &31.5 &42.7 &36.2  \\
       \cmidrule{2-12}
      Relation net~\cite{sung2018learning}& V & Neural network &93.4 &30.0 &45.3 &61.1 &38.1 &47.0 &- &- &-  \\
      \cmidrule{2-12}
      MCZSL~\cite{verma2021meta}& V & Neural network &77.9 &67.1 &72.1 &57.2 & 66.4 &61.4 &40.3 &46.9 &43.4  \\
      \cmidrule{2-12}
      AREN~\cite{xie2019attentive}& S & Ensemble classifier &79.1 &54.7  &64.7  &69.0 &63.2 &66.0 &32.3 &40.3 &35.9 \\
      \cmidrule{2-12}
      GEM-ZSL~\cite{liu2021goal}& S & Softmax & 77.5 & 64.8 & 70.6 & 77.1 & 64.8 & 70.4 & 35.7 & 38.1 & 36.9 \\
      \cmidrule{2-12}
      SELAR~\cite{yang2020simple}&S & Softmax &78.7  &32.9 &46.4 &76.3 &43.0 &55.0 &37.2 &23.8 &29.0 \\
      \cmidrule{2-12}
      Zhu~\cite{zhu2019semantic}& V & - & 87.1 & 37.6 & 52.5 & 71.3 & 36.7 & 48.5 &- &- & -  \\
      \cmidrule{2-12}
      SCILM~\cite{ji2019semantics} & V & Softmax &77.3 &39.2  &52.1 &- &- &- &- &- &-\\
      \cmidrule{2-12}
      IGSC~\cite{yeh2020zero}&S & Softmax &83.6 &25.7 &39.3 &60.2 &40.8 &48.7 &31.3 &39.4 &34.9 \\
      \cmidrule{2-12}
      JIL~\cite{liu2019joint} & L & - &69.0 &47.4 &56.2 &41.4 &38.6 &40.1 &- &- &- \\
      \cmidrule{2-12}
      DTNet~\cite{ji2020dual}& V & Nearest neighbor &77.3 &56.5 &65.3 &53.5 &44.9 &48.9 &- &- &- \\
      \cmidrule{2-12}
      Annadani and Biswas~\cite{biswas2018preserving} & S \& V & Neural network &73.8 &20.7 &32.3 &54.3 &24.6 &33.9 &37.2 &20.8 &26.7 \\
      \cmidrule{2-12}
      PQZSL~\cite{li2019compressing} & L & Nearest neighbor &70.9 &31.7 &43.8 &51.4 &43.2 &46.9 &35.3 &35.1 &35.2 \\
      \cmidrule{2-12}
      LESAE~\cite{liu2018zero}& S \& V & Nearest neighbor &70.6 &21.8 &33.3 &53.0 &24.3 &33.3 &34.7 &21.9 &26.9 \\
      \cmidrule{2-12}
      MCGM-VAE~\cite{shao2020generalized}&L& Softmax &69.3 &60.9 &64.8 &58.0  &51.1 &54.3 &43.8 &38.6 &41.1\\

      \cmidrule{2-12}
      Cacheux et al.~\cite{Cacheux2019modeling}&S \& L & Neural network &83.2 &48.5  &61.3  &52.3 &55.8 &53.0 &30.4 &47.9 &36.8 \\
      \cmidrule{2-12}
      TCN~\cite{jiang2019transferable} & L & Neural network &65.8 &61.2 &63.4 &52.0 &52.6 &52.3 &37.3 &31.2 &34.0 \\
      \cmidrule{2-12}
      CenterZSL~\cite{jin2019discriminant} & S & Softmax &  78.0&  48.6&  59.9& 65.3 &55.8 &60.1 &30.5 &45.8 &36.6 \\
      \cmidrule{2-12}
      SeeNet-HAF~\cite{jin2019beyond} & S & - &82.0  &55.6  &66.3  &67.9 &62.9 &65.3 &33.8 &46.4 &39.1 \\
       \cmidrule{2-12}
      CDL~\cite{jiang2018learning} & L& Nearest neighbor &- &- &- &55.2 &23.5 &32.9 &- &- &- \\
      \cmidrule{2-12}
      ALS~\cite{li2020ajoint} &L& Nearest neighbor &56.0 &53.8 &54.9 &51.6 &43.1 &46.9 &31.9 &41.5 &36.1 \\
      \cmidrule{2-12}
      APNet~\cite{liu2020attribute} & S & Nearest neighbor &83.9  &54.8 &66.4  &55.9 &48.1 &51.7 &40.6 &35.4 & 37.8  \\
      \cmidrule{2-12}
      DDIP~\cite{wang2020learning} & L & Neural network &65.9 &53.6 &59.1 &37.5 &36.6 &37.1 &27.7 &36.8 &31.6\\
    \bottomrule
     \end{tabular}
    \end{adjustbox}
\end{table*}

Although visual embedding models are able to alleviate the hubness problem~\cite{liu2020label}, they suffer from several issues.
Firstly, the visual features and semantic representations are obtained independently, and they are heterogeneous, e.g., they are from different spaces.
As such, the data distributions of both spaces can be different, in which two close categories in one space can be located far away in other space~\cite{fu2015transductive}.
In addition, regression-based methods can not explicitly discover the intrinsic topological structure between two spaces.
Therefore, learning a projection function directly from the space of one modality to the space of another modality may cause information loss, increased complexity, and overfitting toward the seen classes {\color{black}(see Table~\ref{Table:embeding}).}
To mitigate this issue, several studies have attempted to project visual features and semantic representations into a latent space.
Such a projection can reconcile the structural differences between the two spaces.
Besides that, bidirectional learning models aim to learn a better projection and adjust the seen-unseen class domains.
However, these models still suffer from the projection domain shift and bias problems.\par

\begin{table*}[ht!]  \color{black}
\centering
\caption{\label{Table:gen}\color{black} A summary of generative-based methods (``-" indicates the item is not reported by the corresponding study). }
   \begin{adjustbox}{width=0.99\textwidth}
    \begin{tabular}{l c c c c c c c c c c c c}
    \toprule
   \multirow{2}{*} { Study} & \multirow{2}{*} {Generative model} & \multirow{2}{*} {Classifier} &\multicolumn{3}{c}{AWA2} & \multicolumn{3}{c}{CUB} & \multicolumn{3}{c}{SUN} \\
      \cmidrule{4-12}
      &&&$Acc_s$&$Acc_u$&$H$&$Acc_s$&$Acc_u$&$H$&$Acc_s$&$Acc_u$&$H$\\
      \midrule
      GAZSL~\cite{zhu2018agenerative} & GAN & Nearest neighbor & 86.9 & 35.4 & 50.3 & 61.3 & 31.7 & 41.8 & 39.3 & 22.1 & 28.3  \\
      \cmidrule{2-12}
      SE-GZSL~\cite{kumar2018generalized} & VAE/CGAN & Softmax/SVM & 68.1 & 58.3 & 62.8 & 53.3 & 41.5 & 46.7 & 30.5 & 40.9 & 34.9  \\
       \cmidrule{2-12}
      EXEM~\cite{changpinyo2020classifier} & - & Nearest neighbor & 82.2 & 55.0 & 65.9 & 52.1 & 49.8 & 50.9 & 39.1 & 43.5 & 41.2 \\
      \cmidrule{2-12}
      ZSML~\cite{verma2020meta}& GAN & Softmax & 74.6 & 58.9 & 65.8 & 52.1 & 60.0 & 55.7 &- &- & -  \\
      \cmidrule{2-12}
      E-PGN~\cite{yu2020episode}& GAN & Softmax & 83.5 & 52.6 & 64.6 & 61.1 & 52.0 & 56.2 & - &- & -  \\
      \cmidrule{2-12}
      TGMZ~\cite{liu2021task}& CWGAN & SVM/Softmax & 77.3 & 64.1 & 70.1& 56.8 & 60.3 & 58.5 & - & - & -  \\
      \cmidrule{2-12}
      Meta-VGAN~\cite{verma2021towards}& CWGAN & SVM/Softmax & 70.5 & 57.4 & 63.5 & 48.0 & 55.2 & 53.2 & - &- & -  \\

      \cmidrule{2-12}
      f-CLSWGAN~\cite{xian2018feature} & WGAN & Softmax & 61.4 & 57.9 & 59.6 & 57.7 & 43.7 & 49.7 & 36.6 & 42.6 & 39.4  \\
      \cmidrule{2-12}
      OT-ZSL~\cite{wang2021zero} & WGAN & Softmax & 77.5 & 60.6 & 68.0 & 68.0 & 52.7 & 59.4 & 52.7 & 52.7 & 54.8 \\
       \cmidrule{2-12}
      TFGNSCS~\cite{lin2019transfer} & CWGAN & Softmax & 69.5 & 60.3 & 64.5 & 59.7 & 47.6 & 53.0 & 37.5 & 44.3 & 40.6  \\
       \cmidrule{2-12}
       LisGAN~\cite{li2019leveraging} & CWGAN & Softmax & 76.3 & 52.6 & 62.3 & 57.9 & 46.5 & 51.6 & 37.8 & 42.9 & 40.2 \\
       \cmidrule{2-12}
       SPGAN~\cite{ma2020simila} & CWGAN & Softmax & - & - & - & 45.6 & 58.8 & 51.4 & 44.2 & 36.9 & 40.2 \\
       \cmidrule{2-12}
       SR-GAN~\cite{ye2019sr} & WGAN & - & - & - & - & 60.1 & 31.3 & 41.3 & 38.3 & 22.1 & 27.4  \\
       \cmidrule{2-12}
      MGA-GAN~\cite{xie2021generalized} & CWGAN & Softmax & 67.7 & 59.3 & 63.2 & 58.3 & 46.6 & 51.8 & 37.3 & 45.6 & 41.0  \\
      \cmidrule{2-12}
       MKFNet-NFG~\cite{xiang2021multi} & CGAN & Nearest Neighbour & 92.2 & 61.9 & 74.0 & 64.8 & 58.9 & 61.7 & - & - & -  \\
       \cmidrule{2-12}
        Xie et al.\cite{xie2021cross} & CWGAN & Nearest Neighbour & 92.6 & 61.2 & 73.7 & 50.2 & 57.8 & 53.7 & - & - & - \\
       \cmidrule{2-12}
       DASCN~\cite{jian2019dual}& WGAN & Softmax & 68.0 & 59.3 & 63.4 & 59.0 & 45.9 & 51.6 & 38.5 & 42.4 & 40.3 \\

       \cmidrule{2-12}
        cycle-(U)WGAN~\cite{felix2018multi}& WGAN & Softmax & 63.4 & 59.6 & 59.8 & 59.3 & 47.9 & 53.0 & 33.8 & 47.2 & 39.4  \\
       \cmidrule{2-12}
       AFC-GAN~\cite{li2019alleviating}& WGAN & Softmax & - & - & - & 59.7 & 53.5 & 56.4 & 36.1 & 49.1 & 41.6  \\
       \cmidrule{2-12}
       Li et al.~\cite{li2021investigating} & CWGAN & Neural network & - & - & - & 58.6 & 52.3 & 55.3 & 35.1 & 49.3 & 41.0  \\
       \cmidrule{2-12}
        TACO~\cite{chandhok2021two} & CWGAN & Softmax & 74.2 & 59.4 & 66.0 & 60.0 & 51.8 & 55.6 & - & - & - \\
        \cmidrule{2-12}
        CVAE-ZSL~\cite{mishra2018generative} & CVAE & SVM & - & - & 51.2 & - & - & 34.5 & - & - & 26.7 \\
        \cmidrule{2-12}
        Zero-VAE-GAN~\cite{gao2020zerovaegan} & VAE/CGAN & Softmax & 71.7 & 56.2 & 63.0 & 48.5 & 41.1 & 44.4 & 30.9 & 44.4 & 36.5  \\
       \cmidrule{2-12}
       GDAN~\cite{huang2019generative} & CVAE & Neural network & 67.5 & 32.1 & 43.5 & 66.7 & 39.3 & 49.5 & 89.9 & 38.1 & 53.4  \\
       \cmidrule{2-12}
       CADA-VAE~\cite{schonfeld2019generalized} & VAE & Softmax & 75.0 & 55.8 & 63.9 & 53.5 & 51.6 & 52.4 & 35.7 & 47.2 & 40.6  \\
       \cmidrule{2-12}
       Chen et al.~\cite{chen2021entropy} & VAE & Softmax & 78.9 & 55.2 & 64.9 & 55.1 & 50.8 & 52.9 & 36.8 & 44.1 & 40.1  \\
       \cmidrule{2-12}
        Li et al.~\cite{li2019generalized} & MLP & Neural network & 60.9 & 52.4 & 56.3 & 63.4 & 30.2 & 40.9 & 59.0 & 32.2 & 41.6  \\
       \cmidrule{2-12}
       Gao et al.~\cite{gao2018joint} & CVAE/CGAN & Softmax & 71.7 & 56.2 & 63.0 & 45.6 & 42.7 & 44.1 & 30.9 & 44.4& 36.5  \\
       \cmidrule{2-12}
       Xu et al.~\cite{xu2021dual} & CVAE & Softmax & 76.4 & 60.1 & 67.3 & 61.9 & 53.8 & 57.6 & 37.4 & 48.3 & 42.1  \\
       \cmidrule{2-12}


       SAN~\cite{tang2020san}& GAN & Neural network &80.4 & 57.6 &67.10 &49.4 & 48.6  & 49.0 &37.2  & 45.6 & 41.0  \\
       \cmidrule{2-12}
       LIUF~\cite{li2020learning} & MLP & Softmax & 83.5 & 60.6 & 70.2 & 54.0 & 51.2 & 52.5 & 40.4 & 45.7 & 42.9  \\
       \cmidrule{2-12}

       SE-GAN~\cite{pambala2020generative} & WGAN & Integrated Classifier & 79.3 & 55.9 & 65.5 & 55.4 & 52.4 & 53.8 & 51.3 & 34.7 & 41.4 \\

       \cmidrule{2-12}

       BAAE~\cite{yu2018bi} & AE & Softmax/KNN/SVM & 85.6 & 51.4 & 64.22 & - & - & - & 36.7 & 23.1 & 28.4  \\
       \cmidrule{2-12}
      MAAE~\cite{ji2019multi}& AE & Softmax/KNN/SVM & 85.6 & 51.4 & 64.2 & - & - & - & 36.7 & 23.1 & 28.4 \\
        \cmidrule{2-12}



    GDFN.~\cite{luo2020generative} & CVAE & Neareat Neighbour & 85.6 & 38.1 & 52.7 & 57.2 & 40.6 & 47.5 & 35.9 & 39.2 & 39.0  \\
     \cmidrule{2-12}

    Feng and Zhao~\cite{feng2020transfer} & mutli-modal & DAP-based & 72.1 & 63.9 & 67.7 & 42.2 & 44.8 & 43.5 & - & - & -\\

        \bottomrule
     \end{tabular}
   \end{adjustbox}
\end{table*}

{\color{black}\textbf{\textit{Inductive setting vs. Transductive setting}}: As stated earlier, GZSL methods under the transductive setting know the distribution of the unseen classes, as they have access to the unlabeled samples of the unseen classes.
Therefore, they can solve the bias and shift problems.
This results in a balanced $Acc_s$ and $Acc_u$ performance with a high $H$ score as compared with those from GZSL methods under the inductive and semantic transductive settings (Table~\ref{Table:trans} \textit{vs.} Tables~\ref{Table:embeding} and~\ref{Table:gen}).
}

\begin{table*}[tb]  \color{black}
\centering
\caption{\label{Table:trans} A summary of transductive-based methods (``S", ``V" and ``L" represent the semantic, visual and latent embedding spaces, respectively, and ``-" indicates the score is not reported by the corresponding study). }
   \begin{adjustbox}{width=1\textwidth}
    \begin{tabular}{l c c c c c c c c c c c c}
    \toprule
   \multirow{2}{*} { Study} & \multirow{2}{*} {Embedding/generative} & \multirow{2}{*} {Classifier} &\multicolumn{3}{c}{AWA2} & \multicolumn{3}{c}{CUB} & \multicolumn{3}{c}{SUN} \\
      \cmidrule{4-12}
      &&&$Acc_s$&$Acc_u$&$H$&$Acc_s$&$Acc_u$&$H$&$Acc_s$&$Acc_u$&$H$\\
      \midrule

SABR-T~\cite{paul2019semantically}& L & Softmax & 91.0 & 79.7 & 85.0 & 73.7 & 67.2 & 70.3 & 41.5 & 58.8 & 48.6 \\
\cmidrule{2-12}
ERPL~\cite{guan2018extreme}& V& Nearest neighbor & 66.4 & 50.5 & 57.4 & 43.7 & 37.2 & 40.2 & - & - & - \\
\cmidrule{2-12}
Hu et al.~\cite{huo2018zero} & V & Nearest neighbor & 67.8 & 58.7 & 62.9 & 41.6 & 38.0 & 39.7 & - & - & - \\
\cmidrule{2-12}
Long et al.~\cite{long2018pseudo}& L & Nearest neighbor & - & - & - & 51.6 & 23.0 & 31.8 & 32.7 & 19.0 & 24.0 \\
\cmidrule{2-12}
Zhang et al.~\cite{zhang2020towards}& L & Nearest neighbor & 93.2 & 68.4 & 78.9 & 62.9 & 54.0 & 58.1 & 38.5 & 47.2 & 42.4 \\
\cmidrule{2-12}
AEZSL~\cite{niu2018zero} & S & - & - & - & - & - & - & 42.1 & - & - & 36.5\\
\cmidrule{2-12}
Zero-VAE-GAN.~\cite{gao2020zerovaegan}& VAEGAN & Nearest neighbor & 87.0 & 70.2 & 77.6 & 57.9 & 64.1 & 60.8 & 35.8 & 53.1 & 42.8 \\
\cmidrule{2-12}
QFSL~\cite{song2018transductive} & S  & Softmax & 93.1 & 66.2 & 77.4 & 74.9 & 71.5 & 73.2 & 31.2 & 51.3 & 38.8  \\
\cmidrule{2-12}

DTN~\cite{zhang2020deep}& V & Softmax & - & - & - & 66.0 & 42.6 & 51.8 & 38.7 & 35.8 & 37.2  \\
\cmidrule{2-12}
VAEGAN-D2~\cite{xian2019f}& VAEGAN & Softmax & 88.6 & 84.2 & 86.7 & 65.1 & 61.4 & 63.2 & 41.9 & 60.6 & 49.6 \\
\cmidrule{2-12}
SDGN~\cite{wu2020self}& CGAN & Softmax & 89.3 & 88.8 & 89.1 & 70.2 & 69.9 & 70.1 & 46.0 & 62.0 & 52.8 \\

    \bottomrule
     \end{tabular}
   \end{adjustbox}
\end{table*}

Generative-based methods are more effective in solving the bias problem, owing to synthesizing visual features for the unseen classes.
{\color{black}This leads to a slightly balanced $Acc_s$ and $Acc_u$ performance, and consequently higher harmonic mean ($H$), as compared with those from embedding based methods (see Tables~\ref{Table:embeding} and \ref{Table:gen}).
However, the generative based methods are susceptible to the bias problem. }
While, the availability of visual samples for the unseen classes allows the models to perform recognition of both seen and unseen classes in a single process, they generate visual features through learning a model class conditioned on the seen classes.
They do not learn a generic model for generalization toward both seen and unseen class generations~\cite{verma2020meta}.
They are complex in structure and difficult to train (owing to instability).
Their performance is restricted either by obtaining the distribution of visual features using semantic representations or using the Euclidean distance as the constraint to retain the information between the generated visual features and real semantic representations.
In addition, the unconstrained generation of data samples for the unseen classes may produce samples far from their actual distribution.
These methods have access to the semantic representations (under semantic transductive setting) and unlabeled unseen data (under transductive setting), which violate the GZSL setting.\par

\vspace{-0.25cm}
\subsection{Research Gaps}
\label{Sec:sec:future}
Despite considerable progresses in GZSL, methodologies, there are challenges pertaining to the unavailability of visual samples for the unseen classes.  Based on our findings, the main research gaps that require further investigations are as follows:

\begin{itemize}
\item \textbf{Methodology:} Although many studies to solve the projection domain shift problem are available in the literature, it remains a key challenging issue with respect to the existing models.
Most of the existing GZSL methods are based on ideal data sets, which is unrealistic in real-world scenarios.
In practice, the ideal setting is affected by uncertain disturbances, e.g. a few samples in each category are different from other samples.
Therefore, developing robust GZSL models is crucial.
To achieve this, new frameworks that incorporate domain classification without relying on latent space learning are required.\par

On the other hand, techniques that are capable of solving supervised classification problems can be adopted to solve the GZSL problem, e.g. ensemble models~\cite{hansen1990neural} and meta-learning strategy~\cite{verma2020meta}. 
Ensemble models employ a number of individual classifiers to produce multiple predictions, and the final decision is reached by combining the predictions~\cite{pourpanah2019animproved,pourpanah2016aqlearning}.
Recently, Felix et al.~\cite{felix2019multi} introduced an ensemble of visual and semantic classifiers to explore the multi-modality aspect of GZSL. Meta-learning aims to improve the learning ability of the model based on the experience of several learning episodes~\cite{liu2020attribute,verma2020meta}.
GZSL also can be combined with reinforcement learning~\cite{sutton1998introduction,gowda2021claster,pourpanah2019reinforced} to better tackle new tasks.
In addition, GZSL can be extended in several fronts, which include multi-modal learning~\cite{parida2020coordinated,mazumder2021avgzslnet,felix2020augmentation}, multi-label learning~\cite{gupta2021generative}, multi-view learning~\cite{Paul2021generalized}, weakly supervised learning to progressively incorporate training instances from easy to hard~\cite{yu2018zero,xu2021semi}, continual learning~\cite{verma2021meta}, long-tail learning~\cite{samuel2021generalized}, or online learning for few-shot learning where a small portion of labeled samples from some classes are available~\cite{kumar2018generalized}.\par

{\color{black} Separating the seen class domain from unseen class domain using outlier detection (or OoD) techniques is interesting, and these techniques have shown promising results in solving GZSL tasks. However, there are several potential issues. Firstly, this approach requires training several models, i.e., an outlier detection method to separate seen classes from unseen ones, a supervised learning model to categorize seen class samples and a ZSL model to recognize unseen class samples. These requirements lead to computationally expensive and time-consuming issues. Secondly, separating the seen classes from unseen ones using a novelty detector is a difficult task. As the prior information of the unseen classes is usually unknown (e.g. inductive and semantic transcutive settings) and there may be overlapping regions between the seen and unseen class samples, as well as difficulty in tuning model parameters, it is possible for the novelty detector to misclassify part of the data set. This issue is more challenging for fine-grained classes, where they contain discriminative information only in a few regions. Thirdly, visual or semantic features may not be discriminative enough to separate seen classes from unseen ones. Thus, combining information of both visual and semantic spaces is imperative. These issues and the bias problem can result in incorrect final predictions. To alleviate these challenges, Dong et al.~\cite{dong2018learning} categorized the test images into the compositional spaces, including source, target and uncertain spaces.  The uncertain space contains test samples that cannot be confidently classified as one of the source and target spaces. Then, statistical methods are used to analyse the instance distribution and classify the ambiguous instances. However, this approach requires further investigation to study its effectiveness under various settings.}

{\color{black}Recently, transformer-based language models have shown superior performance on various NLP tasks. As an example, generative pre-training (GPT)~\cite{radford2018improving} is a semi-supervised technique that combines unsupervised pre-training with supervised fine-tuning techniques. It uses a large corpus of unlabelled text along with a limited number of manually annotated samples in its operation. GPT-2~\cite{radford2019language} and GPT-3~\cite{brown2020language} are improved models of GPT, which are trained on large-scale data sets for solving tasks under ZSL and few-shot settings, respectively. Moreover, CLIP~\cite{radford2021learning} and DALL-E~\cite{ramesh2021zero} are transformer-based techniques for zero-shot text-to-image generation. However, this field requires further investigations.}
{\color{black}Their ability in learning directly from raw texts (millions of webpages) pertaining to images without any explicit supervision can be adopted for solving GZSL tasks. Unlike attributes defined by humans that contain limited information about the classes, transformers can be leveraged to identify more precise mapping functions between seen and unseen classes in the latent space. Their strong ability in learning from large scale training data sets constitutes a good direction for further research.}\par

\item \textbf{Data:} Semantic representations play an important role in bridging the gap between the seen and unseen classes.
The existing data sets use human-defined attributes or word vectors.
The former shares the same attributes for different classes, and human labor is required for annotation, which is not suitable for large-scale data sets. The latter automatically extracts information from a large test corpus, which is usually noisy.
Therefore, extracting useful knowledge from such data sets is difficult, especially fine-grained data sets~\cite{parida2020coordinated,mazumder2021avgzslnet}.
In this regard, using other modalities such as audio can improve the quality of the data samples. As such, it is crucial to focus on developing new techniques to automatically generate discriminative semantic attribute vectors, exploring other semantic spaces or combinations of various semantic embeddings that can accurately formulate the relationship between seen and unseen classes, and consequently solve the projection domain shift and bias problems.
In addition, there are many unforeseen scenarios in real-world applications, such as driver-less cars or action recognition that may not be included among the seen classes.
While GZSL can be applied to tackle such problems, no suitable data sets for such applications are available at the moment.
This constitutes a key focus area for future research.
 \end{itemize}

\subsection{Concluding Remarks}
\label{Sec:con}
Building models that can simultaneously perform recognition for both seen (source) and unseen (target) classes are vital for advancing intelligent data-based learning models.
This paper, to the best of our knowledge, for the first time presents a comprehensive review of GZSL methods.
Specifically, a hierarchical categorization of GZSL methods along with their representative models has been provided.
In addition, evaluation protocols including benchmark problems and performance indicators, together with future research directions have been presented.
This review aims to promote the perception related to GZSL.

\ifCLASSOPTIONcaptionsoff
  \newpage
\fi

\vspace{-0.5cm}
\bibliographystyle{IEEEtran}
\bibliography{mybib}
\vspace{-1.5cm}

\end{document}